\documentclass{dcogpaper}

\usepackage{fix-cm}
\usepackage{float}
\usepackage{tikz}
\usetikzlibrary{arrows.meta,calc,fit,positioning}

\newtheorem{theorem}{Theorem}
\newtheorem{proposition}{Proposition}
\newtheorem*{informalprop}{Informal Proposition}
\newtheorem{corollary}{Corollary}

\DeclareMathOperator*{\argmin}{arg\,min}
\newcommand{\E}{\mathbb E}
\newcommand{\inner}[2]{\left\langle #1,#2\right\rangle}
\newcommand{\norm}[1]{\left\lVert #1\right\rVert}
\newcommand{\pos}[1]{\left[#1\right]_{+}}

\title{Screen Before You Add: Reducing Candidate Inference in Model Ensembles
}
\author{Davide Mambelli\thanks{Correspondence: [first name]@distributed.inc.}\and Alessandro Rozza\texorpdfstring{\footnotemark[1]}{}\and Steven Motta\texorpdfstring{\footnotemark[1]}{}}
\affiliation{distributed cognition}
\date{2026-10-14}
\shorttitle{Screen Before You Add}
\paperstatus{Preprint}
\paperlogo{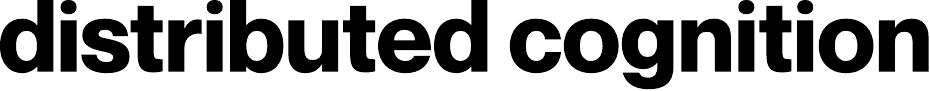}
\papersymbol{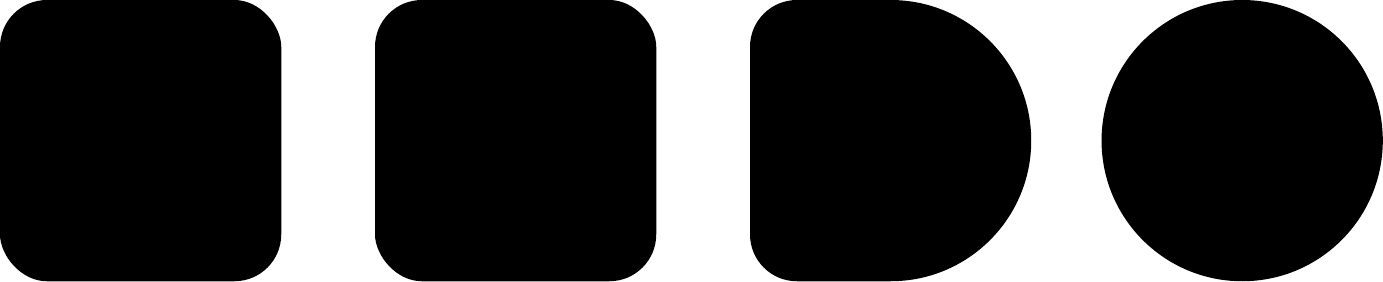}
\papercopyright{\textcopyright\ 2026 distributed cognition. All rights reserved.}
\begin{document}
\maketitle

\begin{abstract}
As new models become available, existing ensembles must be updated to be competitive, making them intrinsically dynamic. Deciding whether to add a candidate model can require inference on an entire annotated dataset. Such a screening becomes prohibitive for models with costly inference especially when they ultimately add no value. Even under simple convex aggregation, relying on benchmark scores can be misleading as strong standalone performance does not guarantee ensemble improvement, and fitted ensemble weights can hurt test performance.
In this work, we study candidate screening and show how early rejection can be achieved safely, exploiting residual geometry for stratification and rejecting the candidate when the best possible remaining predictions cannot achieve the necessary improvement.
The key result that justifies this approach is that, under convex aggregation and squared or Brier loss, the population optimum improves if and only if a candidate induces a prediction change positively aligned with the residuals of the population-optimal existing ensemble.
An optimistic quadratic program (QP) provides certified early rejection, while posterior-scenario sampling offers an alternative heuristic.
We evaluate the approach by showing the tradeoff between saved calls and screening accuracy, which refers to recovering the same accept/reject result as for the entire annotated dataset. We assess empirically whether full-dataset decisions predict candidate usefulness on a held-out test set. On two LLM panels, stratified screening with certified QP rejection saves 21–23\% of candidate calls while keeping 100\% screening accuracy. In our experiments, sampling posterior scenarios further increases call savings to approximately 37\%, with a loss of less than one percentage point in screening accuracy.
\end{abstract}

\section{Introduction}

Ensemble methods combine predictions from multiple models to improve
accuracy by exploiting differences in their errors
\citep{breiman1996stacked,reeve2023diversity}.
Combining capabilities from different models is also an important element
of modern LLM systems \citep{chen2025harnessing}, this is achieved via direct aggregation, as in EVA and
PackLLM \citep{xu2024eva,mavromatis2024packllm}, or model coordination,
as in Fugu \citep{sakanaai2026fugu}. Routing methods such as RouteLLM
instead select a model for each input to reduce inference costs while
maintaining response quality \citep{ong2025routellm}.

Maintaining such model pools raises the recurring \textbf{screening} question of whether a newly available model is worth adding.
The rapid growth in foundation-model releases makes this question increasingly relevant \citep{maslej2024aiindex}.
We call the ensemble formed from the existing models the \emph{incumbent
ensemble}, borrowing terminology from AutoML, where ``incumbent'' denotes
the best solution found so far \citep{hutter2011sequential}.
We use a \emph{screening set} of queries, annotations, and cached
incumbent predictions to fit aggregation weights and compare losses
with and without a candidate. Only candidate predictions require new
inference, but full evaluation requires querying the entire set.
For expensive models, this can make the screening decision costly
even when the candidate ultimately adds no value.

Standalone performance does not settle this question. A candidate's value
depends on the errors left by the existing ensemble and on how the models
work together. A weaker model can correct some of those errors, while a
stronger model can be redundant. Under convex aggregation, the refitted
ensemble can assign the candidate zero weight, so its optimized loss on the
screening set cannot increase. Nevertheless, it may result in no gain and weights fitted
on the screening set can perform worse on held-out examples.
Table~\ref{tab:llm-roster} illustrates this across twelve LLMs and
15 task families. Qwen3 improves held-out accuracy by a mean of
1.47 percentage points on seven tasks, but hurts four and leaves four
unchanged. Every tested model hurts at least one task, including the largest and best standalone performer. An oracle that
already knows which tasks benefit in held-out accuracy can skip
all candidate calls on the others, avoiding 40.0--93.3\% of
screening-set calls across models.

We study screening under convex aggregation with squared or multiclass Brier loss \citep{brier1950verification}.
Adding a candidate allows all ensemble weights to change.
This setting enables an exact residual-alignment characterization of population improvement and a convex QP for safe early rejection on a finite screening set.
Convex aggregation under squared loss is an established setting in the theoretical literature \citep{lecue2013empirical}.
Practical precedents for this setup include MSE-optimized LLM ensembles for well-being score prediction \citep{ravenda2025evidence} and Brier-optimized combinations of LLM forecasts with prediction-market probabilities \citep{alur2025aia}.

\begin{figure}[t]
    \centering
    \resizebox{\textwidth}{!}{% Extra-compact Figure 1, designed at the 13.97 cm ICLR text width.
% All labels, vectors, 24 query cells and the 10/24 stopping point are retained.
% Requires tikz with arrows.meta and calc; no additional packages.
\begingroup%
\definecolor{FigInk}{HTML}{24313D}%
\definecolor{FigMuted}{HTML}{667482}%
\definecolor{FigRule}{HTML}{D4DCE3}%
\definecolor{FigGreen}{HTML}{26744B}%
\definecolor{DraftInk}{HTML}{24313D}%
\definecolor{DraftMuted}{HTML}{667482}%
\definecolor{DraftRule}{HTML}{CDD5DC}%
\definecolor{DraftBlue}{HTML}{246799}%
\begin{tikzpicture}[
    x=1cm,y=1cm,
    every node/.style={text=FigInk,font=\sffamily\fontsize{6.6}{7.6}\selectfont,inner sep=1pt},
    heading/.style={anchor=west,font=\sffamily\bfseries\fontsize{7.8}{9}\selectfont},
    detail/.style={font=\sffamily\fontsize{6.2}{7.2}\selectfont},
    label/.style={anchor=west,font=\sffamily\bfseries\fontsize{6.8}{8}\selectfont},
    note/.style={text=DraftMuted,font=\sffamily\fontsize{6.2}{7.2}\selectfont},
    arrow/.style={draw=DraftMuted,line width=0.6pt,-{Latex[length=1.4mm,width=1mm]}}
]
\path[use as bounding box] (0,0) rectangle (13.97,3.90);
\node[heading] at (0.05,3.70) {(a) Why a candidate helps};
\node[heading] at (4.59,3.70) {(b) Reject early, save candidate calls};
\draw[FigRule,line width=0.5pt] (4.35,0.12) -- (4.35,3.81);

% (a) Same residual geometry, with nearby labels and a shared lower baseline.
\coordinate (origin) at (0.40,1.15);
\coordinate (residualtip) at (0.40,2.86);
\coordinate (helpfultip) at (3.55,2.55);
\coordinate (sidetip) at (3.77,1.15);
\coordinate (projection) at (0.40,2.55);
\draw[FigInk,line width=0.8pt,-{Latex[length=1.8mm,width=1.2mm]}]
    (origin) -- (residualtip);
\draw[FigGreen,line width=1.7pt] (origin) -- (projection);
\draw[FigGreen,line width=1.15pt,-{Latex[length=1.8mm,width=1.2mm]}]
    (origin) -- (helpfultip);
\draw[FigGreen,densely dashed,line width=0.55pt]
    (projection) -- (helpfultip);
\draw[FigMuted,densely dashed,line width=0.75pt,
    -{Latex[length=1.6mm,width=1.1mm]}] (origin) -- (sidetip);
\draw[FigMuted,line width=0.5pt]
    ($(origin)+(0.16,0)$) -- ($(origin)+(0.16,0.16)$)
    -- ($(origin)+(0,0.16)$);
\fill[FigInk] (origin) circle (1.3pt);
\node[anchor=north east,inner sep=1pt] at (origin) {$f_A^\star$};
\node[anchor=west,align=left] at (0.64,3.16)
    {Remaining error\\$r_A=\mu-f_A^\star$};
\node[anchor=north,text=FigGreen,align=center,
    font=\sffamily\fontsize{6}{6.8}\selectfont] at (3.12,1.92)
    {Candidate direction\\$d_b=b-f_A^\star$};
\node[anchor=west,text=FigGreen,align=left,detail] at (0.57,2.14)
    {Corrective\\component};
\node[anchor=north,text=FigMuted,align=center,
    font=\sffamily\fontsize{5.8}{6.6}\selectfont] at (2.085,1.04)
    {Disagreement with\\no correction};
\node[anchor=south,align=center] at (2.14,0.15)
    {$b$ improves $\Longleftrightarrow\ \langle r_A,d_b\rangle>0$};

% (b) Same four strata and the same ten observed predictions among 24 queries.
\node[label] at (4.59,3.13) {Stratified queries};
\foreach \row in {0,...,3} {
    \foreach \col in {0,...,5} {
        \draw[draw=DraftRule,fill=white,line width=0.55pt]
            ({4.65+0.42*\col},{2.54-0.42*\row}) rectangle ++(0.30,0.30);
    }
}
\foreach \row/\col in {0/1,0/4,1/0,1/3,2/0,2/2,2/5,3/1,3/3,3/4} {
    \fill[DraftBlue]
        ({4.65+0.42*\col},{2.54-0.42*\row}) rectangle ++(0.30,0.30);
}
\fill[DraftBlue] (4.65,0.93) rectangle ++(0.14,0.14);
\node[anchor=west,note] at (4.90,1.00) {Queried};
\draw[DraftRule,line width=0.55pt] (4.65,0.55) rectangle ++(0.14,0.14);
\node[anchor=west,note] at (4.90,0.62) {Skipped after rejection};
\draw[arrow] (7.23,2.05) -- (7.85,2.05);

% The horizontal positions are exact call fractions: 4/24, 7/24 and 10/24.
\def\BoundStart{8.09}
\def\BoundFull{13.74}
\pgfmathsetmacro{\BoundFirst}{\BoundStart+4/24*(\BoundFull-\BoundStart)}
\pgfmathsetmacro{\BoundSecond}{\BoundStart+7/24*(\BoundFull-\BoundStart)}
\pgfmathsetmacro{\BoundStop}{\BoundStart+10/24*(\BoundFull-\BoundStart)}
\pgfmathsetmacro{\SavedMid}{(\BoundStop+\BoundFull)/2}
\node[label] at (8.06,3.13) {Upper bound on possible gain};
\draw[DraftMuted,line width=0.5pt] (\BoundStart,1.15) -- (\BoundStart,2.87);
\draw[arrow] (\BoundStart,1.15) -- (13.87,1.15);
\draw[DraftMuted,densely dashed,line width=0.5pt]
    (\BoundStart,1.91) -- (\BoundFull,1.91);
\node[anchor=south east,note] at (\BoundFull,2.00) {Required gain};
\draw[DraftBlue,line width=1.2pt]
    (\BoundStart,2.70) -- (\BoundFirst,2.70) -- (\BoundFirst,2.38)
    -- (\BoundSecond,2.38) -- (\BoundSecond,2.06)
    -- (\BoundStop,2.06) -- (\BoundStop,1.60);
\draw[DraftBlue,densely dotted,line width=0.55pt]
    (\BoundStop,1.60) -- (\BoundStop,1.15);
\fill[DraftBlue] (\BoundStop,1.60) circle (1.5pt);
\node[anchor=west,text=DraftBlue,font=\sffamily\bfseries\fontsize{6.8}{8}\selectfont]
    at ({\BoundStop+0.17},1.58) {Reject};
\node[anchor=north,note] at (\BoundStart,1.00) {0};
\node[anchor=north east,note] at (\BoundFull,1.00) {Full screening set};
\node[anchor=west,note] at (\BoundStart,0.50) {Candidate calls};
\draw[DraftMuted,line width=0.55pt]
    (\BoundStop,0.67) -- (\BoundStop,0.51)
    -- (\BoundFull,0.51) -- (\BoundFull,0.67);
\node[anchor=south,font=\sffamily\bfseries\fontsize{6.8}{8}\selectfont]
    at (\SavedMid,0.15) {Calls avoided};
\end{tikzpicture}%
\endgroup%
}
    \caption{\textbf{Candidate screening from cached predictions.}
    (a) Under squared loss, positive residual alignment characterizes
    improvement of the population-optimal convex ensemble.
    (b) Rows represent strata; filled cells are queried and empty cells
    are skipped. Rejection is certified when the optimistic gain bound
    falls below the required improvement.}
    \label{fig:method-overview}
\end{figure}
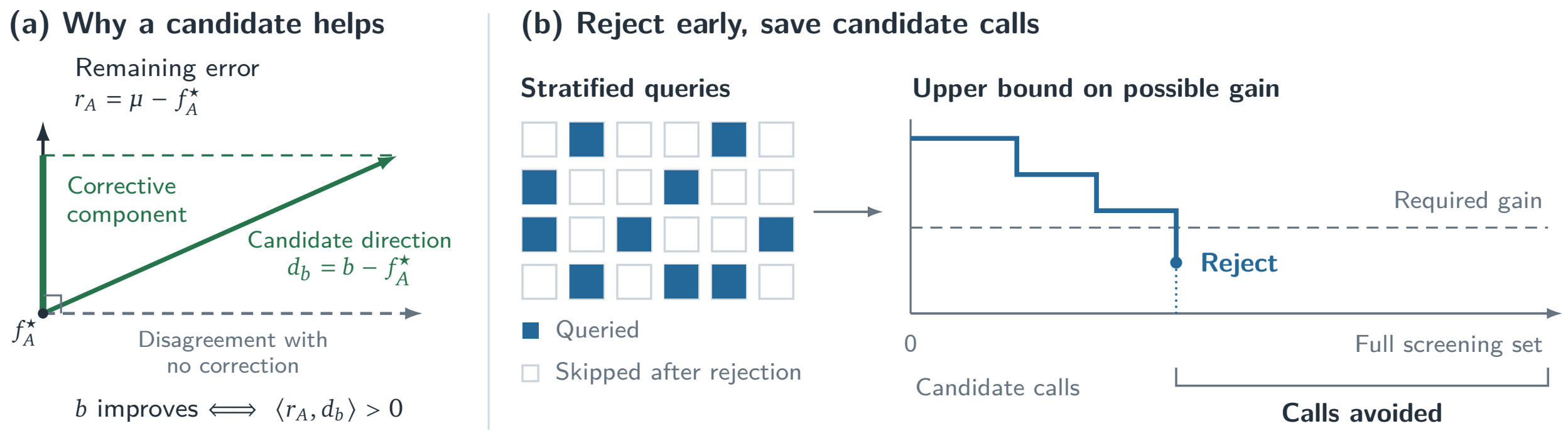

Our goal is to recover the full-screening-set keep/reject decision at a 1\% relative-loss improvement threshold using fewer candidate calls.
Cached residuals and prediction bounds guide stratification, grouping queries by uncertainty in their alignment contributions (Figure~\ref{fig:method-overview}).
An optimistic quadratic program certifies rejection when even the most favorable unobserved predictions cannot achieve the required gain.
A posterior-scenario extension models these predictions to trade some screening accuracy for additional call savings.
We measure calls saved and \textit{screening accuracy}: the percentage of keep/reject decisions matching evaluation on the full screening set.

\begin{table}[!t]
    \definecolor{AccuracyGain}{HTML}{26744B}
    \definecolor{AccuracyLoss}{HTML}{B3261E}
    \centering
    \fontsize{9.5}{11.5}\selectfont
    \setlength{\tabcolsep}{1.5pt}
    \caption{\textbf{Effects of adding each model on 15 tasks.} Held-out top-1 accuracy; weights minimize screening-set Brier loss. Helped/hurt/unchanged give task counts; means give within-group ensemble accuracy changes (pp). Standalone accuracy and calls saved are in \%. A perfect oracle skips harmful and unchanged tasks at zero decision cost. Details and size conventions: Appendix~\ref{app:llm-setup}.}
    \begin{tabular}{@{}l*{8}{c}|*{4}{c}@{}}
        \toprule
        & \multicolumn{8}{c|}{\shortstack{Original models (Each added to the other 7)}} & \multicolumn{4}{c}{\shortstack{New models (Added to the original 8)}} \\
        \cmidrule(lr){2-9}\cmidrule(lr){10-13}
         & {\fontsize{8.5}{9.5}\selectfont Qwen3} & {\fontsize{8.5}{9.5}\selectfont \shortstack{Llama\\3.2}} & {\fontsize{8.5}{9.5}\selectfont \shortstack{Gemma\\4}} & {\fontsize{8.5}{9.5}\selectfont \shortstack{Ministral\\3}} & {\fontsize{8.5}{9.5}\selectfont \shortstack{Phi-4\\Mini}} & {\fontsize{8.5}{9.5}\selectfont \shortstack{Granite\\4.1}} & {\fontsize{8.5}{9.5}\selectfont \shortstack{OLMo\\3}} & {\fontsize{8.5}{9.5}\selectfont \shortstack{Falcon-\\H1}} & {\fontsize{8.5}{9.5}\selectfont SmolLM3} & {\fontsize{8.5}{9.5}\selectfont \shortstack{MiniCPM\\4.1}} & {\fontsize{8.5}{9.5}\selectfont GLM-4} & {\fontsize{8.5}{9.5}\selectfont Apertus} \\
        \midrule
        Size & 4B & 3B & E2B & 3B & 3.8B & 3B & 7B & 7B & 3B & 8B & 9B & 8B \\
        Standalone & 78.14 & 65.60 & 68.35 & 69.73 & 71.21 & 73.62 & 70.10 & 79.98 & 67.34 & 74.16 & 79.91 & 59.17 \\
        Helped & 7 & 2 & 4 & 4 & 2 & 5 & 5 & 9 & 4 & 5 & 8 & 1 \\
        Mean gain & \textcolor{AccuracyGain}{$+1.47$} & \textcolor{AccuracyGain}{$+1.41$} & \textcolor{AccuracyGain}{$+0.94$} & \textcolor{AccuracyGain}{$+1.41$} & \textcolor{AccuracyGain}{$+0.47$} & \textcolor{AccuracyGain}{$+0.79$} & \textcolor{AccuracyGain}{$+0.75$} & \textcolor{AccuracyGain}{$+3.38$} & \textcolor{AccuracyGain}{$+0.94$} & \textcolor{AccuracyGain}{$+1.46$} & \textcolor{AccuracyGain}{$+2.45$} & \textcolor{AccuracyGain}{$+0.31$} \\
        Hurt  & 4 & 5 & 1 & 5 & 2 & 2 & 2 & 3 & 2 & 3 & 2 & 2 \\
        Mean loss & \textcolor{AccuracyLoss}{$-1.02$} & \textcolor{AccuracyLoss}{$-0.46$} & \textcolor{AccuracyLoss}{$-1.56$} & \textcolor{AccuracyLoss}{$-0.50$} & \textcolor{AccuracyLoss}{$-0.42$} & \textcolor{AccuracyLoss}{$-0.47$} & \textcolor{AccuracyLoss}{$-0.94$} & \textcolor{AccuracyLoss}{$-0.52$} & \textcolor{AccuracyLoss}{$-0.47$} & \textcolor{AccuracyLoss}{$-0.42$} & \textcolor{AccuracyLoss}{$-0.47$} & \textcolor{AccuracyLoss}{$-0.78$} \\
        Unchanged & 4 & 8 & 10 & 6 & 11 & 8 & 8 & 3 & 9 & 7 & 5 & 12 \\
        Calls saved & 53.3 & 86.7 & 73.3 & 73.3 & 86.7 & 66.7 & 66.7 & 40.0 & 73.3 & 66.7 & 46.7 & 93.3 \\
        \bottomrule
    \end{tabular}
    \label{tab:llm-roster}
\end{table}

Experiments cover 35 regression tasks and two LLM panels.
On the LLM panels, certified screening saves 21--23\% of candidate
calls relative to full evaluation while maintaining 100\% screening
accuracy. Posterior scenarios raise savings to about 37\% while
retaining over 99\% screening accuracy. We separately check whether
full-set decisions predict improvement on held-out examples.

Our contributions are:
\textbf{(i)} an exact characterization of candidate improvement under
population-optimal convex aggregation with squared or Brier loss;
\textbf{(ii)} a screening method combining residual-based stratification
with certified early rejection to reduce candidate calls; and
\textbf{(iii)} a posterior-scenario extension, evaluated alongside
certified screening on regression and LLM tasks to measure the tradeoff
between screening accuracy and call savings.

\section{Related work}
\label{sec:related-work}

\paragraph{Aggregation.}
Stacking learns how to combine predictors
\citep{wolpert1992stacked,breiman1996stacked}, and statistical aggregation
studies excess risk for model selection, linear combinations, and convex
combinations \citep{tsybakov2003optimal}. These lines of research set the foundations for modern aggregation techniques answering \textbf{how} to aggregate models. In our work, we ask how sample-efficiently we can answer \textbf{when} one supplied candidate can improve an existing convex ensemble.

\paragraph{Residual correction.}
Our screening approach is based on residual geometry. Residuals are at the core of methods like gradient boosting, where they guide how to construct new components \citep{friedman2001greedy}. In our setting,
the candidate already exists and evaluating its predictions is costly.
Prior work on ensemble diversity relates disagreement to individual errors and
the aggregation rule \citep{brown2005diversity,reeve2023diversity}.
Our projection criterion states exactly when this combination of quality
and disagreement permits an improvement.

\paragraph{Sample-efficient evaluation.}
Active testing selects examples to estimate risk with fewer labels
\citep{kossen2021active}. Compact LLM benchmarks use selected subsets and
psychometric structure to approximate full-benchmark performance
\citep{maiapolo2024tiny}, while surrogate uncertainty can guide approximate
Neyman allocation \citep{liu2026active}. The main distinction with our work is the evaluation target and the available prior information. In our setting, labels and incumbent
predictions are available, while the missing information is the candidate's
output. This is motivated by the practical scenario where a practitioner has already trained an aggregation but new models are released.
We use classical stratified estimation and allocation
\citep{cochran1977sampling} to estimate its residual alignment. The theory
controls this estimate, and the experiments assess agreement with the
ensemble decision from the full screening set.

\paragraph{Safe screening.}
Safe screening uses optimization bounds to discard options without changing the target solution \citep{ndiaye2015gapsafe}.
\citet{dantas2019stable} make elimination safe under incomplete knowledge of the predictor vectors identifying inactive variables in $\ell_1$-regularized regression.
\citet{hutter2009paramils} leverage the logic of stopping when even the best remaining outcome cannot change the decision, so the shared principle with our approach is bounding what the remaining evaluation could achieve.
In our setting, rejection must remain valid after full refitting of ensemble weights, so we bound both the effect of unobserved candidate predictions and the best improvement achievable by the aggregation weights.
Statistical bounds for early termination provide an alternative to considering every admissible completion, while achieving looser statistical guarantees. \citet{maron1993hoeffding} use confidence bounds to reject candidates before completing their evaluation, and confidence sequences provide bounds that remain valid across repeated checks as new data arrive \citep{howard2021confidence}. We take inspiration to use probabilistic evidence and design a posterior-scenario heuristic modeling unobserved candidates to trade off safety for earlier rejection.

\section{Problem Setup and Theory}
\label{sec:geometry}

We first characterize when a candidate improves the best ensemble formed by the incumbent models.
The result captures the candidate’s value after population-level weight reoptimization.

Let $A=\{a_1,\ldots,a_m\}$ be fixed predictors, $b$ be a candidate, and $(X,Y)\sim P$ and $P_X$ denotes the marginal distribution of $X$.
Assume squared loss, $\E[Y^2]<\infty$, $a_j,b\in L^2(P_X)$, and unregularized convex aggregation. Here, $L^2(P_X)$ is the space of functions $f$ with finite expected squared magnitude, $\E[f(X)^2]<\infty$.
For $w\in\Delta_m$, an aggregate is
\begin{equation}
    f_{A,w}=\sum_jw_ja_j,
    \qquad
    \Delta_m=\{w\in\mathbb R^m:w\ge0,\ \mathbf1^\top w=1\}.
    \label{eq:convex-aggregate}
\end{equation}
Let $\mathcal C_A=\operatorname{conv}(A)$ and $\mathcal C_{A+b}=\operatorname{conv}(A\cup\{b\})$, where $\operatorname{conv}(A)$ is the set of all convex combinations of predictors in $A$.
For any predictor $f\in L^2(P_X)$, define $R(f)=\E[(Y-f(X))^2]$ and the corresponding population minimizers by
\begin{equation}
    f_A^\star=\argmin_{f\in\mathcal C_A}R(f),
    \qquad
    f_{A+b}^\star=\argmin_{f\in\mathcal C_{A+b}}R(f).
    \label{eq:hull}
\end{equation}
The candidate's population value is
\begin{equation}
    \Delta_b^\star=R(f_A^\star)-R(f_{A+b}^\star)\ge0.
    \label{eq:population-value}
\end{equation}
Their risk difference measures the improvement available in the expanded convex hull.
When weights are learned from finite data, held-out performance also depends on estimation error (Appendix~\ref{app:population-details}).

Under squared loss, this comparison has a geometric form. We write $\mu(x)=\E[Y\mid X=x]$
and use $\inner{u}{v}=\E[u(X)v(X)]$ with its induced norm. Since
\begin{equation}
    R(f)=\E[\operatorname{Var}(Y\mid X)]+\norm{\mu-f}^2,
    \label{eq:risk-decomposition}
\end{equation}
$f_A^\star$ is the $L^2(P_X)$ projection of $\mu$ onto $\mathcal C_A$, which is the convex combination of incumbent predictors closest to $\mu$ in expected squared distance, $\E[(\mu(X)-f(X))^2]$.
We define the residual, candidate direction, and alignment by
\begin{equation}
    r_A=\mu-f_A^\star,
    \qquad
    d_b=b-f_A^\star,
    \qquad
    c_b=\inner{r_A}{d_b}.
    \label{eq:alignment}
\end{equation}
The projection optimality gives
\begin{equation}
    \inner{r_A}{a-f_A^\star}\le0
    \qquad\text{for every } a\in\mathcal C_A.
    \label{eq:projection-optimality}
\end{equation}
Reweighting the existing models cannot further reduce population risk.
The next theorem characterizes when adding $b$ can.
\begin{theorem}[Exact residual-alignment criterion]
Under population-optimal convex aggregation with squared loss,
\begin{equation}
    \Delta_b^\star>0
    \quad\Longleftrightarrow\quad
    c_b=\inner{r_A}{b-f_A^\star}>0.
    \label{eq:alignment-criterion}
\end{equation}
\end{theorem}

\begin{proof}
If $c_b>0$, the feasible segment $f_\lambda=f_A^\star+\lambda d_b$ has gain
\begin{equation}
    R(f_A^\star)-R(f_\lambda)=2\lambda c_b-\lambda^2\norm{d_b}^2>0
\end{equation}
for sufficiently small $\lambda>0$. Conversely, if $c_b\le0$, every
$g=(1-\lambda)a+\lambda b\in\mathcal C_{A+b}$, with
$a\in\mathcal C_A$ and $\lambda\in[0,1]$, satisfies
\begin{equation}
    \inner{r_A}{g-f_A^\star}=(1-\lambda)\inner{r_A}{a-f_A^\star}+\lambda c_b\le0.
\end{equation}
Thus $f_A^\star$ satisfies the projection condition on the expanded hull and
remains optimal.
\end{proof}

The theorem identifies which part of a candidate's disagreement can improve
the ensemble: its component along the residual of the population-optimal
incumbent. Disagreement orthogonal to the incumbent residual cannot improve the population-optimal ensemble.
When alignment is positive, reoptimizing all weights can improve on the direct
move toward the candidate. When alignment is nonpositive, reoptimization
cannot produce a gain.

\paragraph{Labels, standalone quality, and disagreement.} The criterion involves the unknown conditional mean $\mu$. To connect it to observable data, we express the same alignment using labels. Conditional expectation gives
\begin{equation}
    c_b=\E[(Y-f_A^\star(X))(b(X)-f_A^\star(X))].
    \label{eq:label-alignment}
\end{equation}
The noise term has zero expectation conditional on $X$, so the score does not
require the unknown regression function explicitly.
The population-optimal incumbent predictor $f_A^\star$ and the expectation still have to be estimated in practice.
The same score also connects standalone quality with disagreement. To highlight this we rewrite $c_b$ using
$q_b=\norm{b-f_A^\star}^2$ and expanding $R(b)$ which gives
\begin{equation}
    c_b=\frac{R(f_A^\star)-R(b)+q_b}{2}.
    \label{eq:quality-disagreement}
\end{equation}
This identity explains why standalone quality by itself cannot determine a
candidate's value. In particular, $0<c_b<q_b/2$ describes a candidate that
is worse alone than the incumbent aggregate but improves it at a smaller
convex weight. Appendix~\ref{app:population-details} derives the optimal gain
on the direct segment.

\paragraph{Multiclass Brier loss.}
The argument relies on squared-error geometry, which is also preserved by
multiclass Brier loss for probability predictions. Let
$Y\in\{1,\ldots,K\}$ and let $p:\mathcal X\to\Delta_K$ denote a
probability predictor, where $\Delta_K$ is the $K$-class probability simplex.
Define
\begin{equation}
    R(p)=\E[\|e_Y-p(X)\|_2^2],
    \label{eq:brier-risk}
\end{equation}
where $e_Y\in\mathbb R^K$ is the one-hot vector corresponding to class $Y$.
For square-integrable vector-valued functions, use the inner product
$\inner{u}{v}=\E[u(X)^\top v(X)]$.
The same projection argument then gives
\begin{equation}
    R(p_A^\star)>R(p_{A+b}^\star)
    \quad\Longleftrightarrow\quad
    \E[(e_Y-p_A^\star(X))^\top
    (p_b(X)-p_A^\star(X))]>0,
    \label{eq:brier-alignment}
\end{equation}
where $p_b$ is the candidate probability predictor and $p_S^\star$ denotes
the population-optimal convex aggregate of the probability predictors in
$S$. This is the loss and alignment used in the LLM experiments.

\section{Method}
\label{sec:policy}

The previous section characterized candidate improvement at the population optimum.
We now turn to finite data, aiming to recover the full-screening-set decision with fewer candidate calls.
Using labels and incumbent predictions, we construct a residual-informed stratified query path before observing candidate outputs.
Along this path, an optimistic QP certifies rejection when no completion of the unseen predictions consistent with the known bounds can achieve the required improvement.

\subsection{Screening target and query model}
The screening target is the decision obtained by fitting both ensembles on
the complete screening set. For regression, let $y=(y_1,\ldots,y_n)^\top\in\mathbb R^n$, while for Brier loss, we instead use the concatenated one-hot target vector.
For a screening set of $n$ rows $(x_i,y_i)$, we identify each predictor by its vector with elements $f_i=f(x_i)$ and $b_i=b(x_i)$. We also use the empirical inner product
$\langle u,v\rangle_n=n^{-1}\sum_i u_i^\top v_i$, and its induced norm.
For regression and multiclass Brier loss, respectively, we use
\begin{equation}
    R_n(f)=\frac1n\sum_i(y_i-f_i)^2,
    \qquad
    R_n(p)=\frac1n\sum_i\|e_{y_i}-p_i\|_2^2,
    \label{eq:empirical-risk}
\end{equation}
where $p_i=p(x_i)$.
Let $f_{A,n}^\star$ and $f_{A+b,n}^\star$ minimize empirical risk over the
incumbent and augmented convex hulls. Define
\begin{equation}
    r_{A,n}=y-f_{A,n}^\star,
    \qquad
    c_{b,n}=\inner{r_{A,n}}{b-f_{A,n}^\star}_n,
    \qquad
    \Delta_{b,n}^\star=R_n(f_{A,n}^\star)-R_n(f_{A+b,n}^\star),
    \label{eq:screening-quantities}
\end{equation}
where $n$ indicates fitting on the complete screening set. For the relative-gain criterion we assume $R_n(f_{A,n}^\star)>0$ and call a
candidate material on the screening set when
\begin{equation}
    \Delta_{b,n}^\star\ge\tau_n,
    \qquad\tau_n=\gamma R_n(f_{A,n}^\star),
    \qquad\gamma=0.01.
    \label{eq:screening-threshold}
\end{equation}
This finite-set target is distinct from population improvement.
Whether adding the candidate also achieves the required loss reduction on held-out data is tested separately, not assumed.
Held-out results never guide querying or stopping.
Only the candidate values $b_i$ are initially missing.
For scalar regression, querying row $i$ reveals its prediction and the alignment contribution
\begin{equation}
    z_i=r_{A,n,i}(b_i-f_{A,n,i}^\star).
    \label{eq:alignment-contribution}
\end{equation}
Here, $r_{A,n,i}$ and $f_{A,n,i}^\star$ denote the residual and fitted incumbent ensemble prediction on screening example $i$.

\subsection{Building a candidate-independent query path}
\label{sec:alignment-strata}

Before querying the candidate, we can bound how much each row may contribute to its residual alignment.
Suppose each unseen prediction lies in a known interval $[L_i,U_i]$, its contribution then lies in the envelope $[\ell_i,u_i]$, where
\begin{equation}
    \ell_i=\min\{r_{A,n,i}L_i,r_{A,n,i}U_i\}-r_{A,n,i}f_{A,n,i}^\star,
    \qquad
    u_i=\ell_i+|r_{A,n,i}|(U_i-L_i).
    \label{eq:scalar-envelope}
\end{equation}
For multiclass Brier loss, the probability simplex supplies the prediction
bounds. Set $r_i=e_{y_i}-p_{A,i}^\star$, where $p_{A,i}^\star$ is the $i$th-row of the incumbent probability aggregate fitted on the screening set. The unknown row contribution is
$z_i=r_i^\top(p_{b,i}-p_{A,i}^\star)$, with $p_{b,i}=p_b(x_i)$, whose exact extrema over that simplex are
\begin{equation}
    \ell_i=\min_k r_{ik}-r_i^\top p_{A,i}^\star,
    \qquad
    u_i=\max_k r_{ik}-r_i^\top p_{A,i}^\star,
    \label{eq:brier-envelope}
\end{equation}
where $K_i$ is the number of valid classes for row $i$, then we have $k=1,\ldots,K_i$ and $r_{ik}$ is coordinate $k$ of $r_i$.
One query reveals the complete candidate probability vector on that row.

\paragraph{Allocation and estimation.}
We use these known ranges to allocate queries before seeing any candidate
outputs. We sort rows by envelope width and form four strata as equal in
size as possible. For stratum $h$, let $N_h$ be its size, $W_h=N_h/n$, and
$R_h=\max_{i\in h}u_i-\min_{i\in h}\ell_i$ its outer-envelope range.
The target $\pi_h^\star$ assigns more queries to strata with wider ranges.
We mix it equally with proportional allocation to obtain $\widetilde{\pi}_h$, retaining
a baseline share of queries for every stratum
\begin{align}
    \pi_h^\star=\frac{W_hR_h}{\sum_gW_gR_g},
    \qquad
    \widetilde{\pi}_h=\tfrac12W_h+\tfrac12\pi_h^\star.
    \label{eq:range-optimal-allocation+stabilized-allocation}
\end{align}
Thus, half of the allocation follows stratum size, while the other half favors strata with greater possible alignment variation.
If all ranges vanish, set $\pi_h^\star=W_h$ for every stratum $h$.
To approximately follow the target fractions,
we randomly permute rows within each stratum and query one from each
initially. We then repeatedly select the nonexhausted stratum with largest
\begin{equation}
    \frac{\widetilde{\pi}_h^2}{n_h(n_h+1)}
    \label{eq:greedy-priority}
\end{equation}
and query its next permuted row, where $n_h$ is its
current query count. This produces a single query ordering, independent
of candidate outputs, whose first $B$ rows are used at budget $B$.
At budget $B=\sum_hn_h$, the alignment estimate is
\begin{equation}
    \widehat c_B=\sum_hW_h\overline z_h,
    \label{eq:alignment-estimate}
\end{equation}
where $\overline z_h$ is the mean of the queried alignment contributions in stratum $h$.

\begin{informalprop}[Allocation guarantee, see Proposition~\ref{prop:minimax-range-allocation}]
    For equal-sized strata, $\pi^\star$ minimizes the worst-case variance of the alignment estimate under the specified stratum bounds, allowing fractional query counts and assuming no stratum is fully queried. The mixture $\widetilde{\pi}$ keeps the alignment-error bound no larger than that of proportional allocation and at most $\sqrt{2}$ times that of $\pi^\star$, provided the target fractions are realized exactly.
\end{informalprop}

Each sampled row receives weight $N_h/n_h$ in the refit.
These weights correct for unequal sampling across strata.
For any fixed predictor $f$, $\widehat R_B(f)$ is an unbiased estimator of $R_n(f)$.
Appendix~\ref{app:allocation-statements} states the finite-population,
concentration, and weight bounds; Appendix~\ref{app:allocation-diagnostics}
gives their proofs and the integer scheduler argument.

No fixed nonproportional allocation uniformly outperforms uniform sampling when alignment contributions are unrestricted: for any such allocation, there exist contribution patterns for which uniform sampling has lower variance
(Proposition~\ref{prop:no-pointwise-sampling-dominance}).
Our guarantees concern estimation of residual alignment under the specified
stratum bounds. The keep/reject decision instead depends on the loss after
refitting the ensemble weights, so a more accurate alignment estimate does
not automatically imply a more accurate decision. We therefore test
empirically whether the proposed allocation improves agreement with
decisions from the full screening set.

\paragraph{Incumbent loss as an alternative score.}
Incumbent loss gives another way to prioritize rows using information already
in the screening set. The second scheduler uses the known row loss
\begin{equation}
    \eta_i=
    \begin{cases}
        r_{A,n,i}^2 & \text{for scalar squared loss},\\[2pt]
        \|e_{y_i}-p_{A,i}^\star\|_2^2 & \text{for multiclass Brier loss}.
    \end{cases}
    \label{eq:incumbent-row-loss}
\end{equation}
The following relation explains why this simpler score tracks the available
alignment on each row.

\begin{proposition}[Incumbent loss tracks alignment headroom]
\label{prop:loss-alignment-headroom}
For scalar predictions in a common interval $[L,U]$, the width
$w_i=u_i-\ell_i$ satisfies $w_i=(U-L)\sqrt{\eta_i}$.
For the Brier simplex envelope,
\begin{equation}
    u_i=\eta_i,
    \qquad
    \sqrt{\eta_i}\le w_i\le\sqrt{2\eta_i}.
    \label{eq:brier-headroom}
\end{equation}
For binary rows, $\eta_i=w_i^2/2$ exactly.
\end{proposition}

Appendix~\ref{app:loss-headroom-proof} gives the proof. Loss and width
rank regression rows identically. For Brier loss, incumbent loss
specifies the maximum positive alignment and bounds its width. The loss
scheduler sorts by $\eta_i$ and feeds pseudo-intervals $[0,\eta_i]$ into the
same four-stratum allocator. These intervals are scheduling scores, not
alignment bounds. They can change the allocation even when the rankings
agree. The stopping rule always uses genuine prediction bounds.

\subsection{Safe rejection along the query path}
\label{sec:qp-stopping}

Safe rejection asks whether even the most favorable unseen predictions could
make the candidate useful. After a queried prefix, we reject only if no
compatible completion can attain $\tau_n$. For scalar regression, let $d=n$, $y\in\mathbb R^d$, and $\mathbf A\in\mathbb R^{d\times m}$ be the matrix whose
columns contain the incumbent
predictions. Let $\mathcal I_q\subseteq\{1,\ldots,n\}$ denote the indices of
screening examples queried after $q$ candidate calls. The completion box is
\begin{equation}
    \mathcal B_q=\{v\in\mathbb R^d:
      v_i=b_i\ (i\in\mathcal I_q),\quad
      L_i\le v_i\le U_i\ (i\notin\mathcal I_q)\}.
    \label{eq:optimistic-completion-box}
\end{equation}
Assume finite prediction bounds, so the box is compact and contains the true
candidate. We obtain the most optimistic augmented risk by choosing both the
completion and the ensemble weights:
\begin{align}
    \rho_q^{\mathrm{box}}
    ={}&\min_{\alpha,\beta,t}\frac1d\|y-\mathbf A\alpha-t\|_2^2
    \label{eq:optimistic-completion-qp}\\
    \text{subject to }{}&\alpha\ge0,\quad\beta\ge0,\quad
      \mathbf1^\top\alpha+\beta=1,\nonumber\\
    &t_i=\beta b_i&&(i\in\mathcal I_q),\nonumber\\
    &\beta L_i\le t_i\le\beta U_i&&(i\notin\mathcal I_q).\nonumber
\end{align}
$\alpha\in\mathbb R^m$ contains incumbent weights, \(\beta\) is the candidate weight, and $t=\beta v\in\mathbb R^d$ is the weighted candidate completion. The substitution $t=\beta v$ makes the constraints linear, giving a convex
QP with feasible prediction set
$\operatorname{conv}(\{a_1,\ldots,a_m\}\cup\mathcal B_q).$
For a complete vector $v$, write its gain on the screening set as
\begin{equation}
    \Delta_n(v)=R_n(f_{A,n}^\star)
    -\min_{g\in\operatorname{conv}(A\cup\{v\})}R_n(g),
    \label{eq:completion-gain}
\end{equation}
and define
\begin{equation}
    U_q=\max_{v\in\mathcal B_q}\Delta_n(v).
    \label{eq:max-compatible-gain}
\end{equation}

\begin{theorem}[Complete safe rejection for a fixed completion set]
\label{thm:complete-qp-rejection}
The exact optimistic QP satisfies
\begin{equation}
    U_q=R_n(f_{A,n}^\star)-\rho_q^{\mathrm{box}}.
    \label{eq:qp-gain-identity}
\end{equation}
For $\tau_n>0$, rejecting when $U_q<\tau_n$ is safe and complete relative to
$\mathcal B_q$: otherwise some compatible completion has gain at least
$\tau_n$. Every zero-error deterministic rejection rule based only on the
same queried values and completion set therefore has a rejection region
contained in the exact QP's rejection region.
\end{theorem}

\begin{proof}
The perspective formulation minimizes augmented risk jointly over ensemble
weights and $v\in\mathcal B_q$. Subtracting this joint minimum from incumbent
risk proves the identity. Compactness and continuity ensure the maximum
compatible gain is attained. If $U_q<\tau_n$, no compatible candidate is
material; otherwise an attaining material candidate agrees with every
observation, so a zero-error rule cannot reject it.
\end{proof}

As we show next, the exact QP certificate rejects the earliest among all safe rules after the same queries and considers the same possibilities for the unobserved predictions.

\begin{corollary}[Earliest safe rejection on a fixed path]
\label{cor:earliest-safe-qp}
Among rules inspecting the same nested path at the same checkpoints, the
exact QP rejects at the first checkpoint where any zero-error rule using the
declared completion set can reject.
\end{corollary}

Each query shrinks $\mathcal B_q$, so the maximum compatible gain $U_q$ can
only decrease. Applying the theorem at each checkpoint gives the corollary.
Its comparison holds for the same path and checkpoints; it does not establish
that the chosen path is query-optimal. Early acceptance would require a
guarantee for \emph{every} completion and leads to a different max--min problem which we do not study here
(Appendix~\ref{app:one-sided-certificate}). Thus, we rely on the refit at the budget or census for positive acceptance decisions.

\paragraph{Conservative numerical evaluation \& budget fallback.}
A dual lower bound on the optimistic QP loss gives an upper bound on the candidate's best achievable improvement. This allows certified rejection without solving the QP exactly, although rejection may occur later than with the exact solution. Appendix~\ref{app:qp-implementation} describes the numerical safety margin and implementation details.
If no certificate rejects before budget $B$, we refit both ensembles on the queried prefix using post-stratification weights. Assuming positive fitted incumbent loss, we keep the candidate if the relative weighted loss reduction is at least 1\%. This fallback recovers the reference decision when the full screening set is queried, but may disagree at smaller budgets; only early rejection is certified.

\section{Experiments}

\label{sec:experiments}
We evaluate whether screening can reproduce the keep/reject decision from the full screening set while avoiding candidate calls. We first compare sampling rules at fixed budgets, then hold the query path fixed to measure the additional savings from early rejection. On the same three panels, we also test how posterior scenarios trade screening accuracy for further savings. Separately, we assess whether the full-screening-set decision predicts candidate usefulness on held-out data and quantify the computational overhead of the stopping procedures.

\subsection{Experimental setup and evaluation}

\paragraph{Datasets and models.}
We evaluate on 35 regression datasets from OpenML-CTR23
\citep{fischer2023openmlctr23} and 15 language understanding, reasoning, and classification tasks presented in multiple-choice form (see Appendix~\ref{app:llm-setup} for dataset references)
using the same 12 LLMs listed in Table~\ref{tab:llm-roster}.

For regression, the four candidate families are ridge regression,
random forest, histogram gradient boosting, and multilayer perceptrons.
For each dataset and ten training seeds, a candidate from one family
is evaluated against an ensemble built from the other three,
giving $35\times4\times10=1{,}400$ cases.

The LLM evaluation contains two separate tasks. In the \emph{unseen-task}
evaluation, each of the eight original models is added to the other seven
on 15 task families held out from policy development, giving 120 cases.
In the \emph{unseen-model} evaluation, each of four new-family candidates
is added to the fixed original eight on those same tasks, giving
60 cases. Regression screening sets contain 129-3,000 examples, while
each LLM task has 192 screening examples. Both settings use separate
held-out data. Appendix~\ref{app:experimental-setup} lists the tasks,
model configurations, and data splits.

\paragraph{Metrics.}
\label{sec:screening-metrics}
The reference decision evaluates the candidate on the complete screening set and keeps it if refitting the ensemble reduces MSE or Brier loss by at least 1 percent (Equation~\eqref{eq:screening-threshold}).
We measure two quantities: \emph{screening accuracy}, which is the percentage of decisions matching this reference where each task has equal weight, and \emph{calls saved}, which measures the fraction of total candidate calls avoided, rather than the mean percentage saved per task. We average repeated runs within cases and candidate/seed results within tasks.
We report \emph{relative overhead}: stopping computation time as a percentage of estimated inference time saved, using Qwen3 4B one-token timings (Appendix~\ref{app:runtime-break-even}).

\paragraph{Held-out evaluation.}
We separately assess whether the reference decision predicts usefulness
on unseen examples. We freeze the incumbent and augmented weights
fitted on the complete screening set and label a candidate useful
if the frozen augmented ensemble reduces held-out loss by at least one percent relative to the frozen incumbent ensemble.
The reference decisions agree with these held-out labels in
89.29\% of regression cases, 85.00\% of unseen-task cases, and
83.33\% of unseen-model cases.
This check evaluates the usefulness of the full-screening-set reference, as
held-out predictions never enter querying or stopping.
Appendix~\ref{app:heldout-diagnostics} provides further diagnostics.

\paragraph{Comparison protocol.}
We replay cached predictions over ten runs per case, revealing only requested examples.
Budgets reach 2,048 regression or 192 LLM calls, capped by screening-set size.
Sampling comparisons disable stopping; stopping comparisons share the query order, rejection checks, and fallback.
A no-stop baseline isolates savings from early rejection.
Appendix~\ref{app:uncertainty} details the paired bootstrap intervals.

\subsection{Fixed-budget sampling}
\label{sec:fixed-budget-results}

We first ask whether using residual information to choose queries
improves screening accuracy at a fixed inference budget. Early stopping
is disabled to isolate the effect of sampling.

\begin{figure}[!t]
    \centering
    \includegraphics[width=\textwidth]{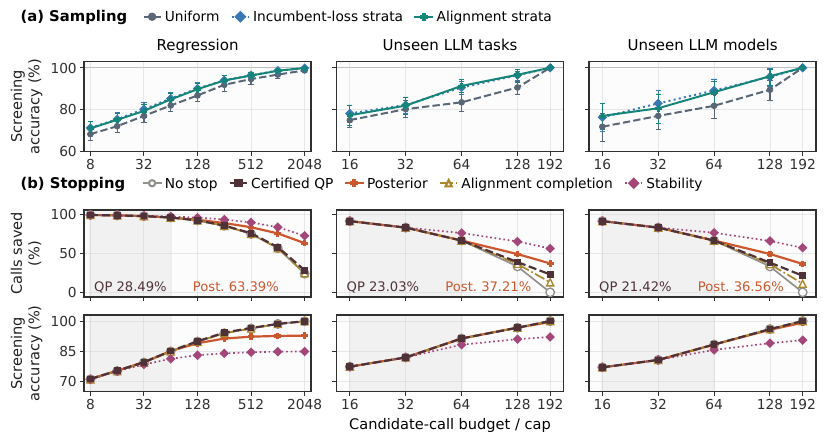}
    \caption{\textbf{(a) Screening accuracy at fixed query budgets.}
    Uniform sampling is compared with incumbent-loss and alignment
    stratification, with early stopping disabled. Bars show pointwise
    95\% hierarchical-bootstrap intervals. Regression budgets are
    capped at each screening set's size. \textbf{ (b) Call savings and screening accuracy.}
    Top: calls saved relative to the complete screening set.
    Bottom: screening accuracy.
    The gray no-stop baseline shows savings from the budget cap alone.
    Posterior scenarios are uncertified and follow QP in the shaded
    region, through 64 calls. Curves show replay point estimates.}
    \label{fig:stratification-and-stopping-by-budget}
\end{figure}

Figure~\ref{fig:stratification-and-stopping-by-budget} (a) shows that both proposed
sampling strategies consistently improve mean screening accuracy
over uniform sampling across all three panels and evaluated query
budgets, with all methods agreeing when the complete screening
set is queried. Neither variant consistently outperforms the other. At 64 candidate calls, the two variants improve screening accuracy
over uniform sampling by 2.9--3.3 percentage points in regression
and 6.5--7.8 points across the two LLM panels.
Residual information therefore helps identify informative queries:
with the same number of candidate calls, screening more reliably
recovers the decision from the complete screening set.

These findings complement the theoretical analysis.
Proposition~\ref{prop:no-pointwise-sampling-dominance} shows that no
fixed nonproportional allocation improves alignment-estimation variance
over uniform sampling for every possible contribution population.
Moreover, estimation guarantees alone do not ensure better decisions
after refitting ensemble weights. Figure~\ref{fig:stratification-and-stopping-by-budget} (a)
provides empirical evidence that both sampling variants help with
this decision on the evaluated tasks.
The variance analysis in
Appendix~\ref{app:exact-variance-limit} further shows that most of
the reduction in alignment-estimation variance comes from allocating
queries unequally across strata, beyond the benefit of forming
strata alone.

\subsection{Certified rejection}
\label{sec:certified-results}

We next ask how many additional calls early rejection can save along the same alignment-stratified query order.
Alongside QP, we compare alignment completion, a simpler certified baseline that bounds each unobserved alignment contribution by its maximum.
It turns this optimistic alignment into an upper bound on the gain after ensemble refitting, whereas QP optimizes jointly over the missing predictions and ensemble weights.
We also compare two-negative stability, a heuristic baseline that rejects after two successive checks indicate insufficient improvement and is motivated by early-stopping criteria in NN training and active learning \citep{prechelt1998early,altschuler2019stopping}.

Figure~\ref{fig:stratification-and-stopping-by-budget}(b) shows that QP saves more calls than alignment completion at the same screening accuracy.
On large budgets, QP saves 21--23\% of candidate calls on the two LLM panels, compared with 11--13\% for alignment completion, while both maintain 100\% screening accuracy.
These budgets cover the complete LLM screening sets, so all savings come from early rejection.
The comparison shows that accounting for ensemble refitting in the optimistic bound enables earlier safe rejection.

In regression, QP saves 28.5\% of calls with 99.89\% screening
accuracy, adding 4.5 percentage points of savings beyond the
24.0\% provided by the budget cap alone.
QP makes no incorrect early rejections on these runs; the remaining
regression disagreements arise from decisions made when the budget
is exhausted. Two-negative stability saves more calls, but incorrectly rejects
candidates that complete evaluation would keep, reducing screening
accuracy.
Certified QP has a relative overhead of only 0.34\%.

\subsection{Posterior relaxation}
\label{sec:posterior-scenario-results}

Certified QP rejection must consider even highly favorable remaining predictions, regardless of how plausible they are given the outputs observed.
The posterior extension uses the candidate predictions already queried, together with the known incumbent predictions and labels, to model the remaining outputs.
We sample complete prediction scenarios, keeping observed outputs fixed, and refit the ensemble weights in each scenario.
Rejection occurs when an upper confidence bound on the modelled probability of achieving the required improvement falls below $\delta$.
This permits earlier rejection at the risk that the actual remaining predictions are more favorable than the model suggests.

At $\delta=0.05$ and the largest budgets, Figure~\ref{fig:stratification-and-stopping-by-budget} (b) shows call savings of approximately 37\% on both LLM panels, with screening accuracy falling by less than one percentage point relative to certified QP.
In regression, savings increase to 63.4\%, while screening accuracy falls from 99.89\% to 92.73\%, showing a larger accuracy cost.
The confidence bound accounts for simulation error under the predictive model; it does not guarantee the rate of incorrect rejections in actual screening.
To avoid basing posterior rejection on too few observed candidate outputs, we enable it only after 64 candidate calls; earlier rejection still uses the certified QP.
Appendix~\ref{app:posterior-scenarios} gives the predictive models, the scenario construction, and the sensitivity to the rejection threshold.
Posterior scenarios increase relative overhead to approximately 3.7\%.

\section{Conclusions}

Our approach reduces the inference needed to decide whether a costly model, such as an LLM, should join an ensemble. The theory and experiments show how information about the existing ensemble can guide this evaluation. Both sampling variants improve mean screening accuracy at partial budgets, while the optimistic QP certifies safe early rejection. Together, they reduce evaluation cost by choosing informative queries and stopping when the candidate cannot achieve the required gain. Full-screening-set decisions also agree substantially with candidate usefulness on held-out data, supporting them as a practical screening target while keeping generalization a separate question.

The early rejection can be conservative because it must consider every admissible set of unobserved predictions, including highly favorable ones. The posterior extension uses observed outputs to model plausible remaining predictions, empirically achieving greater savings at the cost of some incorrect rejections. Future work should establish probabilistic guarantees for these errors under explicit predictive-model assumptions and account for repeated checks. This would let practitioners balance inference savings against an acceptable probability of rejecting a useful candidate.

\section*{Use of AI Tools}
We used generative AI tools to check and assist algebraic proof steps after the authors had developed the theorem statements and proof strategies, to implement the code for numerical evaluation, to produce figures, to edit the text for readability, and to help identify relevant literature. We did not use them to develop theoretical models or conceptual frameworks, to provide critical ingredients of the proofs, to propose or refine hypotheses, to interpret results, or to design or give feedback on the methodology and experiments. Two authors validated the proofs, one author verified the generated code, we checked the generated figures for correctness, all authors reviewed the text, and we verified every paper found through AI-assisted literature search. We take responsibility for the full content of this work, including text, claims, and artifacts produced with the aid of generative AI.

\label{iclr:main-end}

\bibliography{references}

\begin{thebibliography}{60}
\providecommand{\natexlab}[1]{#1}
\providecommand{\url}[1]{\texttt{#1}}
\expandafter\ifx\csname urlstyle\endcsname\relax
  \providecommand{\doi}[1]{doi: #1}\else
  \providecommand{\doi}{doi: \begingroup \urlstyle{rm}\Url}\fi

\bibitem[Altschuler and Bloodgood(2019)]{altschuler2019stopping}
Michael Altschuler and Michael Bloodgood.
\newblock Stopping active learning based on predicted change of {F} measure for text classification.
\newblock In \emph{2019 IEEE 13th International Conference on Semantic Computing (ICSC)}, pages 47--54. IEEE, 2019.
\newblock \doi{10.1109/ICOSC.2019.8665646}.
\newblock URL \url{https://arxiv.org/abs/1901.09118}.

\bibitem[Alur et~al.(2025)Alur, Stadie, Kang, Chen, McManus, Rickert, Lee, Federici, Zhu, Fogerty, Williamson, Lozinski, Linsky, and Sekhon]{alur2025aia}
R.~Alur, B.~C. Stadie, D.~Kang, R.~Chen, M.~McManus, M.~Rickert, T.~Lee, M.~Federici, R.~Zhu, D.~Fogerty, H.~Williamson, N.~Lozinski, A.~Linsky, and J.~S. Sekhon.
\newblock {AIA} forecaster: Technical report.
\newblock \emph{arXiv preprint arXiv:2511.07678}, 2025.
\newblock URL \url{https://arxiv.org/abs/2511.07678}.

\bibitem[Bakouch et~al.(2025)Bakouch, Ben~Allal, Lozhkov, Tazi, Tunstall, Pati{\~n}o, Beeching, Roucher, Reedi, Gallou{\'e}dec, Rasul, Habib, Fourrier, Kydlicek, Penedo, Larcher, Morlon, Srivastav, Lochner, Nguyen, Raffel, von Werra, and Wolf]{bakouch2025smollm3}
Elie Bakouch, Loubna Ben~Allal, Anton Lozhkov, Nouamane Tazi, Lewis Tunstall, Carlos~Miguel Pati{\~n}o, Edward Beeching, Aymeric Roucher, Aksel~Joonas Reedi, Quentin Gallou{\'e}dec, Kashif Rasul, Nathan Habib, Cl{\'e}mentine Fourrier, Hynek Kydlicek, Guilherme Penedo, Hugo Larcher, Mathieu Morlon, Vaibhav Srivastav, Joshua Lochner, Xuan-Son Nguyen, Colin Raffel, Leandro von Werra, and Thomas Wolf.
\newblock {SmolLM3}: smol, multilingual, long-context reasoner.
\newblock Hugging Face technical article, July 2025.
\newblock URL \url{https://huggingface.co/blog/smollm3}.

\bibitem[Breiman(1996)]{breiman1996stacked}
L.~Breiman.
\newblock Stacked regressions.
\newblock \emph{Machine Learning}, 24:\penalty0 49--64, 1996.

\bibitem[Brier(1950)]{brier1950verification}
G.~W. Brier.
\newblock Verification of forecasts expressed in terms of probability.
\newblock \emph{Monthly Weather Review}, 78\penalty0 (1):\penalty0 1--3, 1950.
\newblock \doi{10.1175/1520-0493(1950)078<0001:VOFEIT>2.0.CO;2}.

\bibitem[Brown et~al.(2005)Brown, Wyatt, and Ti\v{n}o]{brown2005diversity}
G.~Brown, J.~Wyatt, and P.~Ti\v{n}o.
\newblock Managing diversity in regression ensembles.
\newblock \emph{Journal of Machine Learning Research}, 6:\penalty0 1621--1650, 2005.

\bibitem[Chen et~al.(2025)Chen, Lu, Li, Chen, Li, Sun, Luo, Mao, Li, Xiao, Yang, Huang, Ban, Sun, and Yu]{chen2025harnessing}
Zhijun Chen, Xiaodong Lu, Jingzheng Li, Pengpeng Chen, Zhuoran Li, Kai Sun, Yuankai Luo, Qianren Mao, Ming Li, Likang Xiao, Dingqi Yang, Xiao Huang, Yikun Ban, Hailong Sun, and Philip~S. Yu.
\newblock Harnessing multiple large language models: A survey on {LLM} ensemble.
\newblock \emph{arXiv preprint arXiv:2502.18036}, 2025.
\newblock URL \url{https://arxiv.org/abs/2502.18036}.

\bibitem[Clopper and Pearson(1934)]{clopper1934use}
C.~J. Clopper and E.~S. Pearson.
\newblock The use of confidence or fiducial limits illustrated in the case of the binomial.
\newblock \emph{Biometrika}, 26\penalty0 (4):\penalty0 404--413, 1934.
\newblock \doi{10.1093/biomet/26.4.404}.

\bibitem[Cochran(1977)]{cochran1977sampling}
W.~G. Cochran.
\newblock \emph{Sampling Techniques}.
\newblock Wiley, third edition, 1977.

\bibitem[Conneau et~al.(2018)Conneau, Rinott, Lample, Williams, Bowman, Schwenk, and Stoyanov]{conneau2018xnli}
A.~Conneau, R.~Rinott, G.~Lample, A.~Williams, S.~R. Bowman, H.~Schwenk, and V.~Stoyanov.
\newblock {XNLI}: Evaluating cross-lingual sentence representations.
\newblock In \emph{Proceedings of EMNLP}, pages 2475--2485, 2018.

\bibitem[Dantas and Gribonval(2019)]{dantas2019stable}
Cassio~Fraga Dantas and R{\'e}mi Gribonval.
\newblock Stable safe screening and structured dictionaries for faster {$\ell_1$} regularization.
\newblock \emph{IEEE Transactions on Signal Processing}, 67\penalty0 (14):\penalty0 3756--3769, 2019.
\newblock \doi{10.1109/TSP.2019.2919404}.
\newblock URL \url{https://arxiv.org/abs/1812.06635}.

\bibitem[Fischer et~al.(2023)Fischer, Harutyunyan, Feurer, and Bischl]{fischer2023openmlctr23}
S.~Fischer, L.~Harutyunyan, M.~Feurer, and B.~Bischl.
\newblock {OpenML-CTR23}: A curated tabular regression benchmarking suite.
\newblock In \emph{AutoML Conference 2023, Workshop Track}, 2023.

\bibitem[Friedman(2001)]{friedman2001greedy}
J.~H. Friedman.
\newblock Greedy function approximation: A gradient boosting machine.
\newblock \emph{The Annals of Statistics}, 29\penalty0 (5):\penalty0 1189--1232, 2001.

\bibitem[{Gemma Team}(2026)]{gemmateam2026gemma4}
{Gemma Team}.
\newblock {Gemma 4} technical report.
\newblock \emph{arXiv preprint arXiv:2607.02770}, 2026.
\newblock URL \url{https://arxiv.org/abs/2607.02770}.

\bibitem[{Granite Team, IBM}(2026)]{ibm2026granite41}
{Granite Team, IBM}.
\newblock {Granite 4.1 LLMs}: How they're built.
\newblock Hugging Face technical article, April 2026.
\newblock URL \url{https://huggingface.co/blog/ibm-granite/granite-4-1}.

\bibitem[Grattafiori et~al.(2024)Grattafiori, Dubey, Jauhri, Pandey, Kadian, Al-Dahle, et~al.]{grattafiori2024llama3}
Aaron Grattafiori, Abhimanyu Dubey, Abhinav Jauhri, Abhinav Pandey, Abhishek Kadian, Ahmad Al-Dahle, et~al.
\newblock The {Llama 3} herd of models.
\newblock \emph{arXiv preprint arXiv:2407.21783}, 2024.
\newblock URL \url{https://arxiv.org/abs/2407.21783}.

\bibitem[Hartvigsen et~al.(2022)Hartvigsen, Gabriel, Palangi, Sap, Ray, and Kamar]{hartvigsen2022toxigen}
T.~Hartvigsen, S.~Gabriel, H.~Palangi, M.~Sap, D.~Ray, and E.~Kamar.
\newblock {ToxiGen}: A large-scale machine-generated dataset for adversarial and implicit hate speech detection.
\newblock arXiv:2203.09509, 2022.

\bibitem[Hoeffding(1963)]{hoeffding1963probability}
W.~Hoeffding.
\newblock Probability inequalities for sums of bounded random variables.
\newblock \emph{Journal of the American Statistical Association}, 58\penalty0 (301):\penalty0 13--30, 1963.

\bibitem[Howard et~al.(2021)Howard, Ramdas, McAuliffe, and Sekhon]{howard2021confidence}
S.~R. Howard, A.~Ramdas, J.~McAuliffe, and J.~Sekhon.
\newblock Time-uniform, nonparametric, nonasymptotic confidence sequences.
\newblock \emph{The Annals of Statistics}, 49\penalty0 (2):\penalty0 1055--1080, 2021.

\bibitem[Hutter et~al.(2011)Hutter, Hoos, and Leyton-Brown]{hutter2011sequential}
F.~Hutter, H.~H. Hoos, and K.~Leyton-Brown.
\newblock Sequential model-based optimization for general algorithm configuration.
\newblock In \emph{Learning and Intelligent Optimization}, volume 6683 of \emph{Lecture Notes in Computer Science}, pages 507--523. Springer, 2011.
\newblock \doi{10.1007/978-3-642-25566-3_40}.
\newblock URL \url{https://www.cs.ubc.ca/~hutter/papers/11-LION5-SMAC.pdf}.

\bibitem[Hutter et~al.(2009)Hutter, Hoos, Leyton-Brown, and St{\"u}tzle]{hutter2009paramils}
Frank Hutter, Holger~H. Hoos, Kevin Leyton-Brown, and Thomas St{\"u}tzle.
\newblock {ParamILS}: An automatic algorithm configuration framework.
\newblock \emph{Journal of Artificial Intelligence Research}, 36:\penalty0 267--306, 2009.
\newblock \doi{10.1613/jair.2861}.
\newblock URL \url{https://arxiv.org/abs/1401.3492}.

\bibitem[Khot et~al.(2020)Khot, Clark, Guerquin, Jansen, and Sabharwal]{khot2020qasc}
T.~Khot, P.~Clark, M.~Guerquin, P.~Jansen, and A.~Sabharwal.
\newblock {QASC}: A dataset for question answering via sentence composition.
\newblock In \emph{Proceedings of AAAI}, pages 8082--8090, 2020.

\bibitem[Klaassen and Lenstra(2008)]{klaassen2008stratification}
C.~A.~J. Klaassen and A.~J. Lenstra.
\newblock Why stratification may hurt, \& how much.
\newblock arXiv:0802.2087, 2008.

\bibitem[Kossen et~al.(2021)Kossen, Farquhar, Gal, and Rainforth]{kossen2021active}
J.~Kossen, S.~Farquhar, Y.~Gal, and T.~Rainforth.
\newblock Active testing: Sample-efficient model evaluation.
\newblock In \emph{Proceedings of the 38th International Conference on Machine Learning}, pages 5753--5763, 2021.

\bibitem[Lai et~al.(2017)Lai, Xie, Liu, Yang, and Hovy]{lai2017race}
G.~Lai, Q.~Xie, H.~Liu, Y.~Yang, and E.~Hovy.
\newblock {RACE}: Large-scale reading comprehension dataset from examinations.
\newblock In \emph{Proceedings of EMNLP}, pages 785--794, 2017.

\bibitem[Lecu\'e(2013)]{lecue2013empirical}
G.~Lecu\'e.
\newblock Empirical risk minimization is optimal for the convex aggregation problem.
\newblock \emph{Bernoulli}, 19\penalty0 (5B):\penalty0 2153--2166, 2013.
\newblock \doi{10.3150/12-BEJ447}.

\bibitem[Li et~al.(2024)]{li2024wmdp}
N.~Li et~al.
\newblock The {WMDP} benchmark: Measuring and reducing malicious use with unlearning.
\newblock In \emph{Proceedings of the 41st International Conference on Machine Learning}, 2024.

\bibitem[Lin et~al.(2022)Lin, Hilton, and Evans]{lin2022truthfulqa}
S.~Lin, J.~Hilton, and O.~Evans.
\newblock {TruthfulQA}: Measuring how models mimic human falsehoods.
\newblock In \emph{Proceedings of ACL}, pages 3214--3252, 2022.

\bibitem[Liu et~al.(2026{\natexlab{a}})Liu, Khandelwal, Subramanian, Jouault, Rastogi, Sad{\'e}, et~al.]{liu2026ministral3}
Alexander~H. Liu, Kartik Khandelwal, Sandeep Subramanian, Victor Jouault, Abhinav Rastogi, Adrien Sad{\'e}, et~al.
\newblock {Ministral 3}.
\newblock \emph{arXiv preprint arXiv:2601.08584}, 2026{\natexlab{a}}.
\newblock URL \url{https://arxiv.org/abs/2601.08584}.

\bibitem[Liu et~al.(2026{\natexlab{b}})Liu, Zhang, Liu, and Zhu]{liu2026active}
Z.~Liu, J.~Zhang, C.~Liu, and Y.~Zhu.
\newblock Active testing of large language models via approximate {Neyman} allocation.
\newblock \emph{arXiv preprint arXiv:2605.10075}, 2026{\natexlab{b}}.

\bibitem[Maas et~al.(2011)Maas, Daly, Pham, Huang, Ng, and Potts]{maas2011sentiment}
A.~L. Maas, R.~E. Daly, P.~T. Pham, D.~Huang, A.~Y. Ng, and C.~Potts.
\newblock Learning word vectors for sentiment analysis.
\newblock In \emph{Proceedings of ACL-HLT}, pages 142--150, 2011.

\bibitem[Maia~Polo et~al.(2024)Maia~Polo, Weber, Choshen, Sun, Xu, and Yurochkin]{maiapolo2024tiny}
F.~Maia~Polo, L.~Weber, L.~Choshen, Y.~Sun, G.~Xu, and M.~Yurochkin.
\newblock {tinyBenchmarks}: Evaluating {LLMs} with fewer examples.
\newblock In \emph{Proceedings of the 41st International Conference on Machine Learning}, pages 34303--34326, 2024.

\bibitem[Maron and Moore(1993)]{maron1993hoeffding}
Oded Maron and Andrew~W. Moore.
\newblock {Hoeffding} races: Accelerating model selection search for classification and function approximation.
\newblock In \emph{Advances in Neural Information Processing Systems}, volume~6, pages 59--66. Morgan Kaufmann, 1993.
\newblock URL \url{https://proceedings.neurips.cc/paper_files/paper/1993/hash/02a32ad2669e6fe298e607fe7cc0e1a0-Abstract.html}.

\bibitem[Maslej et~al.(2024)Maslej, Fattorini, Perrault, Parli, Reuel, Brynjolfsson, Etchemendy, Ligett, Lyons, Manyika, Niebles, Shoham, Wald, and Clark]{maslej2024aiindex}
N.~Maslej, L.~Fattorini, R.~Perrault, V.~Parli, A.~Reuel, E.~Brynjolfsson, J.~Etchemendy, K.~Ligett, T.~Lyons, J.~Manyika, J.~C. Niebles, Y.~Shoham, R.~Wald, and J.~Clark.
\newblock The {AI Index} 2024 annual report.
\newblock Technical report, AI Index Steering Committee, Institute for Human-Centered AI, Stanford University, Stanford, CA, April 2024.
\newblock URL \url{https://hai.stanford.edu/assets/files/hai_ai-index-report-2024-smaller2.pdf}.

\bibitem[Mavromatis et~al.(2024)Mavromatis, Karypis, and Karypis]{mavromatis2024packllm}
C.~Mavromatis, P.~Karypis, and G.~Karypis.
\newblock Pack of {LLMs}: Model fusion at test-time via perplexity optimization.
\newblock In \emph{First Conference on Language Modeling}, 2024.
\newblock URL \url{https://arxiv.org/abs/2404.11531}.

\bibitem[{Microsoft} et~al.(2025){Microsoft}, Abouelenin, Ashfaq, Atkinson, Awadalla, Bach, et~al.]{microsoft2025phi4mini}
{Microsoft}, Abdelrahman Abouelenin, Atabak Ashfaq, Adam Atkinson, Hany Awadalla, Nguyen Bach, et~al.
\newblock {Phi-4-Mini} technical report: Compact yet powerful multimodal language models via {Mixture-of-LoRAs}.
\newblock \emph{arXiv preprint arXiv:2503.01743}, 2025.
\newblock URL \url{https://arxiv.org/abs/2503.01743}.

\bibitem[{MiniCPM Team}(2025)]{minicpm2025minicpm4}
{MiniCPM Team}.
\newblock {MiniCPM4}: Ultra-efficient {LLMs} on end devices.
\newblock \emph{arXiv preprint arXiv:2506.07900}, 2025.
\newblock URL \url{https://arxiv.org/abs/2506.07900}.

\bibitem[Ndiaye et~al.(2015)Ndiaye, Fercoq, Gramfort, and Salmon]{ndiaye2015gapsafe}
E.~Ndiaye, O.~Fercoq, A.~Gramfort, and J.~Salmon.
\newblock {GAP Safe} screening rules for sparse multi-task and multi-class models.
\newblock In \emph{Advances in Neural Information Processing Systems 28}, 2015.

\bibitem[Ong et~al.(2025)Ong, Almahairi, Wu, Chiang, Wu, Gonzalez, Kadous, and Stoica]{ong2025routellm}
I.~Ong, A.~Almahairi, V.~Wu, W.-L. Chiang, T.~Wu, J.~E. Gonzalez, M.~W. Kadous, and I.~Stoica.
\newblock {RouteLLM}: Learning to route {LLMs} with preference data.
\newblock In \emph{The Thirteenth International Conference on Learning Representations}, 2025.
\newblock URL \url{https://openreview.net/forum?id=8sSqNntaMr}.

\bibitem[Pilehvar and Camacho-Collados(2019)]{pilehvar2019wic}
M.~T. Pilehvar and J.~Camacho-Collados.
\newblock {WiC}: The word-in-context dataset for evaluating context-sensitive meaning representations.
\newblock In \emph{Proceedings of NAACL-HLT}, pages 1267--1273, 2019.

\bibitem[Ponti et~al.(2020)Ponti, Glava\v{s}, Majewska, Liu, Vuli\'c, and Korhonen]{ponti2020xcopa}
E.~M. Ponti, G.~Glava\v{s}, O.~Majewska, Q.~Liu, I.~Vuli\'c, and A.~Korhonen.
\newblock {XCOPA}: A multilingual dataset for causal commonsense reasoning.
\newblock In \emph{Proceedings of EMNLP}, pages 2362--2376, 2020.

\bibitem[Prechelt(1998)]{prechelt1998early}
Lutz Prechelt.
\newblock Early stopping---but when?
\newblock In Genevieve~B. Orr and Klaus-Robert M{\"u}ller, editors, \emph{Neural Networks: Tricks of the Trade}, volume 1524 of \emph{Lecture Notes in Computer Science}, pages 55--69. Springer, 1998.
\newblock \doi{10.1007/3-540-49430-8_3}.
\newblock URL \url{https://doi.org/10.1007/3-540-49430-8_3}.

\bibitem[{Project Apertus} et~al.(2025){Project Apertus}, Hern{\'a}ndez-Cano, H{\"a}gele, Huang, Romanou, Solergibert, et~al.]{apertus2025apertus}
{Project Apertus}, Alejandro Hern{\'a}ndez-Cano, Alexander H{\"a}gele, Allen~Hao Huang, Angelika Romanou, Antoni-Joan Solergibert, et~al.
\newblock {Apertus}: Democratizing open and compliant {LLMs} for global language environments.
\newblock \emph{arXiv preprint arXiv:2509.14233}, 2025.
\newblock URL \url{https://arxiv.org/abs/2509.14233}.

\bibitem[{Qwen Team}(2025)]{qwen2025qwen3}
{Qwen Team}.
\newblock {Qwen3} technical report.
\newblock \emph{arXiv preprint arXiv:2505.09388}, 2025.
\newblock URL \url{https://arxiv.org/abs/2505.09388}.

\bibitem[Ravenda et~al.(2025)Ravenda, Kara-Isitt, Swift, Mira, and Raballo]{ravenda2025evidence}
F.~Ravenda, F.-Z. Kara-Isitt, S.~Swift, A.~Mira, and A.~Raballo.
\newblock From evidence mining to meta-prediction: a gradient of methodologies for task-specific challenges in psychological assessment.
\newblock In \emph{Proceedings of the 10th Workshop on Computational Linguistics and Clinical Psychology ({CLPsych} 2025)}, pages 242--248. Association for Computational Linguistics, 2025.
\newblock \doi{10.18653/v1/2025.clpsych-1.20}.
\newblock URL \url{https://aclanthology.org/2025.clpsych-1.20/}.

\bibitem[Rogers et~al.(2020)Rogers, Kovaleva, Downey, and Rumshisky]{rogers2020quail}
A.~Rogers, O.~Kovaleva, M.~Downey, and A.~Rumshisky.
\newblock Getting closer to {AI} complete question answering: A set of prerequisite real tasks.
\newblock In \emph{Proceedings of AAAI}, 2020.

\bibitem[{Sakana AI}(2026)]{sakanaai2026fugu}
{Sakana AI}.
\newblock {Sakana Fugu} technical report.
\newblock arXiv:2606.21228, 2026.
\newblock URL \url{https://arxiv.org/abs/2606.21228}.

\bibitem[Tafjord et~al.(2019)Tafjord, Gardner, Lin, and Clark]{tafjord2019quartz}
O.~Tafjord, M.~Gardner, K.~Lin, and P.~Clark.
\newblock {QuaRTz}: An open-domain dataset of qualitative relationship questions.
\newblock In \emph{Proceedings of EMNLP-IJCNLP}, pages 5941--5946, 2019.

\bibitem[{Team GLM} et~al.(2024){Team GLM}, Zeng, Xu, Wang, Zhang, Yin, et~al.]{glm2024chatglm}
{Team GLM}, Aohan Zeng, Bin Xu, Bowen Wang, Chenhui Zhang, Da~Yin, et~al.
\newblock {ChatGLM}: A family of large language models from {GLM-130B} to {GLM-4} all tools.
\newblock \emph{arXiv preprint arXiv:2406.12793}, 2024.
\newblock URL \url{https://arxiv.org/abs/2406.12793}.

\bibitem[{Team Olmo} et~al.(2025){Team Olmo}, Ettinger, Bertsch, Kuehl, Graham, Heineman, et~al.]{olmo2025olmo3}
{Team Olmo}, Allyson Ettinger, Amanda Bertsch, Bailey Kuehl, David Graham, David Heineman, et~al.
\newblock {Olmo 3}.
\newblock \emph{arXiv preprint arXiv:2512.13961}, 2025.
\newblock URL \url{https://arxiv.org/abs/2512.13961}.

\bibitem[Tsybakov(2003)]{tsybakov2003optimal}
A.~B. Tsybakov.
\newblock Optimal rates of aggregation.
\newblock In \emph{Proceedings of the 16th Annual Conference on Learning Theory}, pages 303--313, 2003.

\bibitem[Warstadt et~al.(2020)Warstadt, Parrish, Liu, Mohananey, Peng, Wang, and Bowman]{warstadt2020blimp}
A.~Warstadt, A.~Parrish, H.~Liu, A.~Mohananey, W.~Peng, S.-F. Wang, and S.~R. Bowman.
\newblock {BLiMP}: The benchmark of linguistic minimal pairs for {English}.
\newblock \emph{Transactions of the Association for Computational Linguistics}, 8:\penalty0 377--392, 2020.

\bibitem[Weso\l{}owski et~al.(2024)Weso\l{}owski, Wieczorkowski, and W\'ojciak]{wesolowski2024recursive}
J.~Weso\l{}owski, R.~Wieczorkowski, and W.~W\'ojciak.
\newblock Recursive {Neyman} algorithm for optimum sample allocation under box constraints on sample sizes in strata.
\newblock \emph{Survey Methodology}, 50\penalty0 (2):\penalty0 487--511, 2024.

\bibitem[Wolpert(1992)]{wolpert1992stacked}
D.~H. Wolpert.
\newblock Stacked generalization.
\newblock \emph{Neural Networks}, 5\penalty0 (2):\penalty0 241--259, 1992.

\bibitem[Wood et~al.(2023)Wood, Mu, Webb, Reeve, Luj\'an, and Brown]{reeve2023diversity}
D.~Wood, T.~Mu, A.~M. Webb, H.~W.~J. Reeve, M.~Luj\'an, and G.~Brown.
\newblock A unified theory of diversity in ensemble learning.
\newblock \emph{Journal of Machine Learning Research}, 24\penalty0 (359):\penalty0 1--49, 2023.

\bibitem[Xu et~al.(2024)Xu, Lu, and Zhang]{xu2024eva}
Y.~Xu, J.~Lu, and J.~Zhang.
\newblock Bridging the gap between different vocabularies for {LLM} ensemble.
\newblock In \emph{Proceedings of the 2024 Conference of the North American Chapter of the Association for Computational Linguistics: Human Language Technologies (Volume 1: Long Papers)}, pages 7140--7152. Association for Computational Linguistics, 2024.
\newblock \doi{10.18653/v1/2024.naacl-long.395}.
\newblock URL \url{https://aclanthology.org/2024.naacl-long.395/}.

\bibitem[Yang et~al.(2019)Yang, Zhang, Tar, and Baldridge]{yang2019pawsx}
Y.~Yang, Y.~Zhang, C.~Tar, and J.~Baldridge.
\newblock {PAWS-X}: A cross-lingual adversarial dataset for paraphrase identification.
\newblock In \emph{Proceedings of EMNLP-IJCNLP}, 2019.

\bibitem[Yu et~al.(2020)Yu, Jiang, Dong, and Feng]{yu2020reclor}
W.~Yu, Z.~Jiang, Y.~Dong, and J.~Feng.
\newblock {ReClor}: A reading comprehension dataset requiring logical reasoning.
\newblock In \emph{International Conference on Learning Representations}, 2020.

\bibitem[Zhang et~al.(2015)Zhang, Zhao, and LeCun]{zhang2015character}
X.~Zhang, J.~Zhao, and Y.~LeCun.
\newblock Character-level convolutional networks for text classification.
\newblock In \emph{Advances in Neural Information Processing Systems 28}, 2015.

\bibitem[Zuo et~al.(2025)Zuo, Velikanov, Chahed, Belkada, Rhayem, Kunsch, et~al.]{zuo2025falconh1}
Jingwei Zuo, Maksim Velikanov, Ilyas Chahed, Younes Belkada, Dhia~Eddine Rhayem, Guillaume Kunsch, et~al.
\newblock {Falcon-H1}: A family of hybrid-head language models redefining efficiency and performance.
\newblock \emph{arXiv preprint arXiv:2507.22448}, 2025.
\newblock URL \url{https://arxiv.org/abs/2507.22448}.

\end{thebibliography}
\bibliographystyle{plainnat}

\clearpage
\appendix

\section{Population identities and finite-sample fitting}
\label{app:population-details}

The population criterion concerns the best functions in the two convex hulls.
Here we record the corresponding learned-weight identities and the best gain
on the direct candidate segment.

\subsection{Learned weights and held-out risk}

For an aggregation-training sample $D_n=\{(X_i,Y_i)\}_{i=1}^n$, let
$\widehat R_n$ denote the empirical risk from Section~\ref{sec:policy}, now
viewed as a function of the random training sample.
The learned incumbent and augmented predictors are
\begin{align}
    \widehat f_A
    &\in\argmin_{f\in\operatorname{conv}(A)}\widehat R_n(f),
    &
    \widehat f_{A+b}
    &\in\argmin_{f\in\operatorname{conv}(A\cup\{b\})}\widehat R_n(f).
    \label{eq:learned-aggregates}
\end{align}
Conditional on $D_n$, the complete learned improvement is
\begin{equation}
    \Delta_b(D_n)=R(\widehat f_A)-R(\widehat f_{A+b}).
    \label{eq:learned-target}
\end{equation}
This is a held-out or deployment quantity. If
\begin{equation}
    \widehat e_A=Y-\widehat f_A(X),
    \qquad
    h_b=\widehat f_{A+b}(X)-\widehat f_A(X),
\end{equation}
then expanding the two squared errors gives the complete identity
\begin{equation}
    \Delta_b(D_n)
    =2\E[\widehat e_Ah_b]-\E[h_b^2].
    \label{eq:complete-improvement}
\end{equation}
Improvement therefore rewards prediction movement aligned with the remaining
error and penalizes its squared size. The identity is exact, but it is not a
cheap screen because $h_b$ is known only after fitting the augmented aggregate.

The optimized training loss cannot answer the held-out question because the
augmented simplex contains the incumbent simplex and therefore cannot have
higher optimized training loss.

\paragraph{Finite-sample transfer.}
The population criterion concerns the best functions in two convex hulls.
After fitting their weights on a finite screening set, the corresponding held-out
improvement also contains the difference in estimation error. Define
\begin{align}
    \mathcal E_A(D_n)
    &=R(\widehat f_A)-R(f_A^\star),\\
    \mathcal E_{A+b}(D_n)
    &=R(\widehat f_{A+b})-R(f_{A+b}^\star).
    \label{eq:excess-risks}
\end{align}

\begin{equation}
    \Delta_b(D_n)
    =
    \Delta_b^\star
    +\mathcal E_A(D_n)
    -\mathcal E_{A+b}(D_n).
    \label{eq:excess-risk-decomposition}
\end{equation}
This follows by adding and subtracting the two population-optimal risks. Thus
the alignment theorem answers whether the hull contains an improving direction,
not whether a finite-sample fit will transfer. Cross-fitting, validation, or a
separate generalization argument is needed for the latter. Our experiments
therefore treat the labeled screening set as a finite optimization problem and score
transfer on a separate held-out split.

\subsection{The best direct segment gain}

Using $q_b$ from Section~\ref{sec:geometry}, define the gain along the direct
segment $f_\lambda=(1-\lambda)f_A^\star+\lambda b$ by
\begin{equation}
    G_b(\lambda)
    :=R(f_A^\star)-R(f_\lambda)
    =2\lambda c_b-\lambda^2q_b,
    \qquad 0\le\lambda\le1.
    \label{eq:direct-path-gain}
\end{equation}
Consequently,
\begin{equation}
    \Delta_b^\star
    \ge G_b^\star
    :=\max_{0\le\lambda\le1}G_b(\lambda)
    =
    \begin{cases}
        0, & c_b\le0,\\[3pt]
        c_b^2/q_b, & 0<c_b<q_b,\\[3pt]
        2c_b-q_b, & c_b\ge q_b,
    \end{cases}
    \label{eq:direct-path-bound}
\end{equation}
obtained by clipping the unconstrained optimum $c_b/q_b$ to $[0,1]$.
This clipping statement applies when $q_b>0$; if $q_b=0$, then $d_b=0$,
$c_b=0$, and $G_b^\star=0$. The sign is controlled by alignment; the
attainable effect also depends on how much irrelevant movement accompanies it.

\section{Sampling guarantees and supporting results}
\label{app:allocation-statements}

Propositions~\ref{prop:no-pointwise-sampling-dominance}
and~\ref{prop:minimax-range-allocation} apply established sampling limitations
and allocation principles to candidate-alignment screening; we do not claim
the underlying principles as novel.

\subsection{Design variance and the scope of pointwise dominance}

Consider $H\ge2$ fixed strata and a fixed allocation of $B$ queries.
Let $N_h$ be the size of stratum $h$, $W_h=N_h/n$, and let $n_h$ rows be drawn
uniformly without replacement inside it. For the complete candidate
contributions $z_i$, write
\begin{equation}
    \mu_h=\frac1{N_h}\sum_{i\in h}z_i,
    \qquad
    S_h^2=\frac1{N_h-1}\sum_{i\in h}(z_i-\mu_h)^2.
\end{equation}
The post-stratified alignment estimator
$\widehat c_B=\sum_hW_h\overline z_h$ has exact conditional design variance
\citep{cochran1977sampling}
\begin{equation}
    \operatorname{Var}(\widehat c_B\mid z)
    =\sum_hW_h^2
      \left(1-\frac{n_h}{N_h}\right)\frac{S_h^2}{n_h}.
    \label{eq:exact-stratified-variance}
\end{equation}

The following counterexample instantiates the established observation that
stratification can increase variance relative to simple random sampling
\citep{klaassen2008stratification}.

\begin{proposition}[No pointwise dominance over uniform sampling]
\label{prop:no-pointwise-sampling-dominance}
Suppose every stratum contains at least two rows and receives at least one
query. For any nonproportional allocation with $0<B<n$, some unrestricted finite
contribution population has larger exact design variance under that allocation than under a
uniform sample of size $B$.
\end{proposition}

\begin{proof}
A nonproportional allocation undersamples some proper stratum $h$, so
$n_h/B<N_h/n$. Choose a population that varies inside this stratum and equals
its stratum mean everywhere else. Its overall finite-population variance is
\begin{equation}
    S^2=\frac{(N_h-1)S_h^2}{n-1}.
\end{equation}
The ratio between
\eqref{eq:exact-stratified-variance} and the variance of a uniform sample is
\begin{equation}
    \frac{W_h^2B(n-1)}{n_h(N_h-1)}
    \frac{1-n_h/N_h}{1-B/n}>1,
\end{equation}
where the first factor is greater than one because
$n_h/B<W_h$ and $W_h(n-1)/(N_h-1)>1$, while the second is greater than one
because $n_h/N_h<B/n$. Hence any fixed nonuniform allocation loses to uniform
sampling for at least one placement of candidate variation.
\end{proof}

The counterexample places variation wherever the fixed allocation undersamples.
It ranges over arbitrary contribution populations and need not satisfy a
particular set of known row envelopes. It therefore does not preclude
improvements for a restricted envelope class.

\subsection{Exact finite-population allocation and stabilization}

For stratum $h$, set
\begin{equation}
    a_h=\min_{i\in h}\ell_i,
    \qquad
    b_h=\max_{i\in h}u_i,
    \qquad
    R_h=b_h-a_h.
    \label{eq:stratum-outer-range}
\end{equation}
The robust class allows every $z_i$ in stratum $h$ to range over the common
outer interval $[a_h,b_h]$. The largest finite-population variance in this
class is
\begin{equation}
    V_h^{\max}
    =R_h^2\frac{\lfloor N_h^2/4\rfloor}{N_h(N_h-1)}.
    \label{eq:max-stratum-variance}
\end{equation}
Applying capped Neyman allocation
\citep{cochran1977sampling,wesolowski2024recursive} to these worst-case stratum
variances gives, ignoring integer rounding and the required initial coverage,
the exact minimax allocation
\begin{equation}
    n_h^{\mathrm{mm}}
    =\min\!\left\{N_h,\lambda W_h\sqrt{V_h^{\max}}\right\},
    \qquad
    \sum_h n_h^{\mathrm{mm}}=B,
    \label{eq:finite-population-minimax-allocation}
\end{equation}
where $\lambda>0$ is chosen to meet the budget among positive-range strata.
If those strata reach their censuses first, any remaining queries can be placed
arbitrarily in zero-range strata. For a continuous target share $\pi_h>0$, define the associated concentration
criterion
\begin{equation}
    C(\pi)=\sum_h\frac{W_h^2R_h^2}{\pi_h}.
    \label{eq:range-allocation-criterion}
\end{equation}

\begin{proposition}[Minimax range allocation and stabilization]
\label{prop:minimax-range-allocation}
Suppose a budget $B$ assigns $n_h=B\pi_h$ queries to each stratum and samples
uniformly without replacement within strata. Over the robust outer-envelope
class, the worst exact design variance equals
\begin{equation}
    \sup_{z_i\in[a_h,b_h]}
    \operatorname{Var}(\widehat c_B\mid z)
    =\sum_hW_h^2
      \left(1-\frac{n_h}{N_h}\right)
      \frac{V_h^{\max}}{n_h},
    \label{eq:worst-case-design-variance}
\end{equation}
and \eqref{eq:finite-population-minimax-allocation} minimizes this quantity over
continuous allocations. For equal-sized, unsaturated strata this minimizer is
$\pi^\star$, which also minimizes $C(\pi)$. Moreover, the post-stratified
estimator satisfies, for every $\delta\in(0,1)$,
\begin{equation}
    \Pr\!\left(
      |\widehat c_B-c_{b,n}|>
      \sqrt{\frac{\log(2/\delta)}{2B}C(\pi)}
    \right)\le\delta.
    \label{eq:stratified-alignment-concentration}
\end{equation}
The stabilized target $\widetilde{\pi}$ obeys
\begin{equation}
    C(\widetilde{\pi})\le\frac{C(W)+C(\pi^\star)}{2}\le C(W),
    \qquad
    C(\widetilde{\pi})\le2C(\pi^\star).
    \label{eq:stabilized-allocation-guarantees}
\end{equation}
If the target shares are realized exactly, every normalized design weight is
at most $2/B$ and the Kish effective sample size is at least $B/2$.
\end{proposition}

These statements concern a fixed continuous allocation for a scalar estimate.
The envelope class allows each row to take any value in its stratum's common
outer interval; it can be larger than the original collection of row intervals.
The implemented schedule stabilizes and discretizes this allocation to support
both estimation and the downstream refit. The exact moment bounds for its
realized integer counts still follow from the stated sampling design; claims
requiring $n_h=B\widetilde{\pi}_h$ describe the continuous target. Appendix
\ref{app:allocation-diagnostics} proves the proposition and the greedy integer
scheduler's separate surrogate optimality.

\subsection{Allocation proofs and integer scheduling}
\label{app:allocation-diagnostics}

We first prove Proposition~\ref{prop:minimax-range-allocation}. For values in
$[a_h,b_h]$, use the identity
\begin{equation}
    S_h^2
    =\frac{1}{N_h(N_h-1)}
      \sum_{i<j,\ i,j\in h}(z_i-z_j)^2.
\end{equation}
This convex function reaches its maximum over the box at an endpoint vector.
If $k$ of the $N_h$ values use one endpoint, then
\begin{equation}
    S_h^2
    =R_h^2\frac{k(N_h-k)}{N_h(N_h-1)}.
\end{equation}
The product $k(N_h-k)$ is largest at the most even split, which gives
\eqref{eq:max-stratum-variance}. Since the strata vary independently, taking
this maximum in every term of \eqref{eq:exact-stratified-variance} proves
\eqref{eq:worst-case-design-variance}.

Write $A_h=W_h^2V_h^{\max}$. Apart from the allocation-independent term
$-\sum_h A_h/N_h$, minimizing the worst-case variance is equivalent to
minimizing
\begin{equation}
    \sum_h\frac{A_h}{n_h}
    \qquad\text{subject to}\qquad
    \sum_hn_h=B,
    \quad
    n_h\le N_h.
    \label{eq:minimax-allocation-program}
\end{equation}
The KKT
conditions give $n_h\propto\sqrt{A_h}=W_h\sqrt{V_h^{\max}}$ until a stratum
reaches its census, after which the remaining budget is allocated by the same
rule. This is exactly \eqref{eq:finite-population-minimax-allocation}. When
all $N_h$ are equal, the multiplier relating $\sqrt{V_h^{\max}}$ to $R_h$ is
common across strata, so the unsaturated solution is $\pi^\star$.

For the concentration statement, let
$\mu_h=N_h^{-1}\sum_{i\in h}z_i$ be the complete mean in stratum $h$. A
uniform sample without replacement has $\E[\overline z_h]=\mu_h$, so
\begin{equation}
    \E[\widehat c_B]=\sum_hW_h\mu_h=c_{b,n}.
\end{equation}
Hoeffding's comparison theorem for
sampling without replacement gives the same bounded-variable moment bound as
independent sampling with replacement \citep{hoeffding1963probability}. Samples
are independent across strata, hence
\begin{equation}
    \Pr\!\left(|\widehat c_B-c_{b,n}|>t\right)
    \le
    2\exp\!\left\{
      -\frac{2t^2}{\sum_hW_h^2R_h^2/n_h}
    \right\}.
\end{equation}
Substituting $n_h=B\pi_h$ proves
\eqref{eq:stratified-alignment-concentration}.

For any allocation $\pi$ on the simplex, Cauchy--Schwarz gives
\begin{equation}
    C(\pi)
    =\sum_h\frac{(W_hR_h)^2}{\pi_h}
    \ge\left(\sum_hW_hR_h\right)^2,
    \label{eq:range-allocation-cauchy}
\end{equation}
with equality at $\pi=\pi^\star$. Terms with $W_hR_h=0$ contribute zero. The map
$\pi\mapsto C(\pi)$ is convex, so for $\widetilde{\pi}=(W+\pi^\star)/2$,
\begin{equation}
    C(\widetilde{\pi})\le\frac{C(W)+C(\pi^\star)}2\le C(W).
\end{equation}
Also $\widetilde{\pi}_h\ge \pi_h^\star/2$, which gives $C(\widetilde{\pi})\le2C(\pi^\star)$. If
$R_{\rm all}=\max_i u_i-\min_i\ell_i$, then
\begin{equation}
    C(W)=\sum_hW_hR_h^2\le R_{\rm all}^2,
\end{equation}
recovering a bound no larger than the
global uniform-range criterion.

When $n_h=B\widetilde{\pi}_h$ exactly, one sampled row in stratum $h$ has normalized design
weight
\begin{equation}
    \omega_i=\frac{W_h}{n_h}=\frac{W_h}{B\widetilde{\pi}_h}\le\frac2B.
\end{equation}
The weights sum to one, so
\begin{equation}
    \sum_i\omega_i^2\le\frac2B\sum_i\omega_i=\frac2B
\end{equation}
and the Kish effective sample
size is at least $B/2$. This completes the proof.

The implementation must use integer counts and one nested path for every
budget. For a fixed target $\widetilde{\pi}$, define
\begin{equation}
    G_{\widetilde{\pi}}(n_1,\ldots,n_H)=\sum_h\frac{\widetilde{\pi}_h^2}{n_h},
    \qquad 1\le n_h\le N_h.
    \label{eq:discrete-allocation-surrogate}
\end{equation}
Adding one query to stratum $h$ decreases $G_{\widetilde{\pi}}$ by exactly
$\widetilde{\pi}_h^2/[n_h(n_h+1)]$. These marginal decreases are nonincreasing within each
stratum. Selecting the largest available decrease at each step therefore
minimizes $G_{\widetilde{\pi}}$ for every integer budget, subject to the coverage and population
constraints. This is the greedy rule used by the nested scheduler. The
continuous factor-two and weight statements describe its target; the variance
diagnostic below uses the realized integer counts instead of treating them as
exactly $B\widetilde{\pi}_h$.

\subsection{Exact design variance and a limitation}
\label{app:exact-variance-limit}

Equation~\eqref{eq:exact-stratified-variance} is available only after the
candidate outputs are known, so we use it as a post-hoc mechanism diagnostic.
Replacing the exact maximum in \eqref{eq:max-stratum-variance} by the simpler
upper bound $N_hR_h^2/[4(N_h-1)]$ gives
\begin{equation}
    \operatorname{Var}(\widehat c_B\mid z)
    \le\sum_hW_h^2
      \left(1-\frac{n_h}{N_h}\right)
      \frac{N_hR_h^2}{4(N_h-1)n_h}.
    \label{eq:finite-population-range-bound}
\end{equation}
Both the exact variance and this bound vanish at a census. Proposition
\ref{prop:no-pointwise-sampling-dominance} says the realized ratio cannot favor
the residual-aware allocation for every finite contribution population. The diagnostic
below asks how often it does so on the frozen panels and how closely the robust
bound tracks the realized variance.

\begin{figure}[!htbp]
    \centering
    \includegraphics[width=\textwidth]{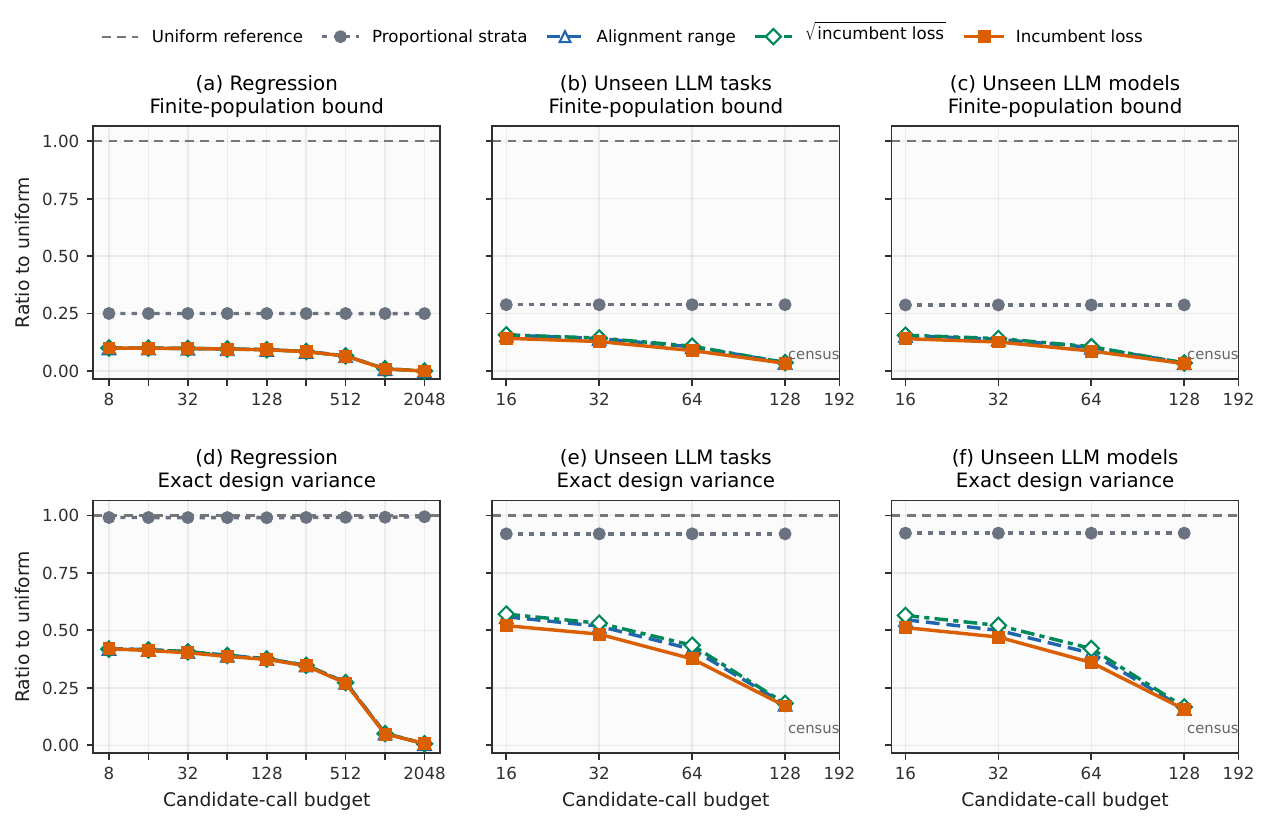}
    \caption{\textbf{Mechanism diagnostic for residual-aware sampling.}
    Columns show regression, unseen LLM task families, and unseen LLM model
    families. Top: the finite-population range bound in
    \eqref{eq:finite-population-range-bound}, divided by the corresponding
    uniform bound. Bottom: exact design variance from
    \eqref{eq:exact-stratified-variance}, divided by uniform variance. Values
    below one favor the residual-aware schedule. QP stopping and convex refits
    are disabled. The 192-row LLM census is omitted because every exact
    variance and finite-population bound is zero.}
    \label{fig:alignment-variance-diagnostics}
\end{figure}

Figure~\ref{fig:alignment-variance-diagnostics} reports the full budget curves.
At 64 calls, loss-based allocation reduces exact variance by 61.4\% in
regression, 62.4\% on unseen LLM tasks, and 64.0\% on unseen LLM models.
Alignment-envelope allocation reduces it by 60.9\%, 58.2\%, and 60.0\%.
Every one of the 35 regression datasets and 15 LLM task families in each panel
has lower aggregated exact variance than uniform sampling. In contrast,
population-proportional allocation inside the same residual strata reduces
pooled exact variance by only 1.0\%, 8.0\%, and 7.7\%. The range-bound ratio is
more optimistic than the realized ratio, especially for proportional strata;
it supports the allocation criterion but does not predict its effect size.

\begin{table}[!ht]
    \centering
    \caption{Exact design variance of the alignment estimator at 64 calls,
    reported as the ratio of summed case variances to the corresponding
    sum under uniform sampling. Lower is better.
    ``Proportional strata'' uses alignment-width strata but allocates queries by
    stratum population alone. The square-root-loss control uses scheduling
    pseudo-intervals $[0,\sqrt{\eta_i}]$. The last column applies to all four
    schedules. Full budget curves and the range-bound diagnostic appear in
    Figure~\ref{fig:alignment-variance-diagnostics} in the appendix.}
    \label{tab:alignment-variance-b64}
    \setlength{\tabcolsep}{3.75pt}
    \begin{tabular}{@{}lrrrrr@{}}
        \toprule
        Panel & Proportional strata & Alignment & $\sqrt{\text{loss}}$ & Loss
              & Tasks below uniform \\
        \midrule
        Regression & 0.990 & 0.391 & 0.390 & \textbf{0.386} & 35/35 \\
        LLMs: unseen tasks & 0.920 & 0.418 & 0.435 & \textbf{0.376} & 15/15 \\
        LLMs: unseen models & 0.923 & 0.400 & 0.421 & \textbf{0.360} & 15/15 \\
        \bottomrule
    \end{tabular}
\end{table}

\begin{samepage}
\subsection{Proof of the loss--headroom relation}
\label{app:loss-headroom-proof}

\begin{proof}
The scalar identity follows immediately from
$u_i-\ell_i=|r_{A,n,i}|(U_i-L_i)$. For Brier loss, suppress the row index, write
$p=p_A^\star$, let $t$ be the true class, and set
\begin{equation}
    x=1-p_t,
    \qquad
    m=\max_{k\ne t}p_k,
    \qquad
    s_2=\sum_{k\ne t}p_k^2.
\end{equation}
The largest alignment contribution is attained by the perfect prediction
$e_t$, so
\begin{equation}
    u=\|e_t-p\|_2^2=x^2+s_2=\eta.
\end{equation}
The largest and smallest entries of the residual vector $e_t-p$ are
$x$ and $-m$, respectively, so the alignment interval has width
$w=x+m$. Since $m^2\le s_2\le xm$,
\begin{equation}
    \frac{w^2}{2}
    \le x^2+s_2
    \le w^2,
\end{equation}
which proves $\sqrt{\eta_i}\le w_i\le\sqrt{2\eta_i}$. With two classes, $m=x$ and
$s_2=x^2$, so equality gives $\eta=w^2/2$.
\end{proof}
\end{samepage}

\section{Supporting completion certificates}
\label{app:completion-details}

\subsection{Why the certificate is one-sided}
\label{app:one-sided-certificate}
A zero-error early-acceptance rule based on the same partial information would
instead require
\begin{equation}
    L_q
    :=\min_{v\in\mathcal B_q}\Delta_n(v)
    =R_n(f_{A,n}^\star)
     -\max_{v\in\mathcal B_q}
       \min_{g\in\operatorname{conv}(A\cup\{v\})}R_n(g)
    \ge\tau_n.
\end{equation}
Unlike $U_q$, this is a max--min problem: the completion is adversarial, while
the ensemble may be reoptimized for that completion. The perspective argument
above therefore does not produce the same convex QP by reversing the
inequality. Interchanging the optimizations gives a safe sufficient condition,
but requires one fixed ensemble weighting to attain the target for every
compatible completion and can be highly conservative for broad completion
sets. We therefore certify early rejection only and reserve positive decisions
for the budget refit or evaluation of the full screening set.

\subsection{Relation to alignment-only screening}

\begin{proposition}[Necessary alignment for material gain on the screening set]
\label{prop:screening-alignment}
For squared loss and simplex aggregation,
\begin{equation}
    \Delta_{b,n}^\star\le2\pos{c_{b,n}}.
    \label{eq:screening-alignment-upper}
\end{equation}
Consequently, $\Delta_{b,n}^\star\ge\tau_n$ requires
$c_{b,n}\ge\tau_n/2$.
\end{proposition}

Appendix~\ref{app:alignment-certificate} proves the proposition and derives its
closed-form partial-query rejection rule. Positive alignment is necessary but
does not guarantee a one-percent gain. The exact QP subsumes this scalar
screen when both use the same completion set. Indeed, write
$c_n(v)=\inner{r_{A,n}}{v-f_{A,n}^\star}_n$. Proposition
\ref{prop:screening-alignment} gives
\begin{equation}
    U_q
    \le 2\max_{v\in\mathcal B_q}\pos{c_n(v)},
    \label{eq:qp-dominates-alignment}
\end{equation}
which is the optimistic alignment-completion bound. Different completion sets
need not preserve this ordering.

\subsection{Alignment-completion rejection}
\label{app:alignment-certificate}

The alignment inequality yields a closed-form rejection certificate even
without solving the QP. It is computationally cheap, but it discards the
quadratic curvature used by the exact optimistic completion. We retain it as a
theoretical comparator rather than as part of the final policy.

\begin{proof}[Proof of Proposition~\ref{prop:screening-alignment}]
Every augmented prediction has the form
$g=(1-\lambda)a+\lambda b$, with $a\in\operatorname{conv}(A)$ and
$\lambda\in[0,1]$. Projection optimality gives
$\inner{r_{A,n}}{a-f_{A,n}^\star}_n\le0$. For
$h=g-f_{A,n}^\star$,
\begin{equation}
    R_n(f_{A,n}^\star)-R_n(g)
    =2\inner{r_{A,n}}{h}_n-\norm{h}_n^2
    \le2\lambda c_{b,n}
    \le2\pos{c_{b,n}}.
\end{equation}
Applying the inequality to $g=f_{A+b,n}^\star$ proves
\eqref{eq:screening-alignment-upper}.
\end{proof}

\begin{corollary}[Alignment-completion rejection]
\label{cor:alignment-completion}
Let $\mathcal I_q$ be the queried rows after $q$ calls. With $z_i$ and $u_i$
defined in Section~\ref{sec:alignment-strata}, set
\begin{equation}
    \overline c_q
    =\frac1n\left(
        \sum_{i\in\mathcal I_q}z_i
        +\sum_{i\notin\mathcal I_q}u_i
    \right).
    \label{eq:completed-alignment}
\end{equation}
If $2\pos{\overline c_q}<\tau_n$, then the candidate cannot achieve the
material gain $\tau_n$ on the screening set.
\end{corollary}

\begin{proof}
Every unseen $z_i$ is at most $u_i$, so $c_{b,n}\le\overline c_q$. Proposition
\ref{prop:screening-alignment} gives
\begin{equation}
    \Delta_{b,n}^\star
    \le2\pos{c_{b,n}}
    \le2\pos{\overline c_q}.
    \label{eq:alignment-completion-upper}
\end{equation}
Thus $2\pos{\overline c_q}<\tau_n$ certifies failure to meet the required gain on the screening set. The same
argument applies to the vector inner products used for multiclass Brier loss.
\end{proof}

\subsection{Weak duality and the numerical witness}
\label{app:weak-dual-certificate}

The following expressions use coordinate-average risk. In scalar regression
$d=n$, so this equals MSE on the screening set. For flattened Brier coordinates multiply
risks and gain bounds by $d/n$ to recover loss per question.

For any $s\in\mathbb R^d$, write $s_{ik}$ for its entry corresponding to
class $k$ of example $i$. For scalar predictions, take $K_i=1$ and identify
$s_{i1}=s_i$, and likewise for candidate predictions and bounds. Define the
support function of the completion box by
\begin{equation}
    h_q(s)
    =\sum_{i\in\mathcal I_q}\sum_{k=1}^{K_i}s_{ik}b_{ik}
     +\sum_{i\notin\mathcal I_q}\sum_{k=1}^{K_i}
       \max\{s_{ik}L_{ik},s_{ik}U_{ik}\},
    \label{eq:completion-support}
\end{equation}
and the explicit weak-dual value
\begin{equation}
    D_q(s)
    =2s^\top y-d\|s\|_2^2
     -2\max\left\{
        \max_{1\le j\le m}a_j^\top s,\ h_q(s)
     \right\}.
    \label{eq:qp-weak-dual}
\end{equation}

\begin{proposition}[Solver-independent weak-dual certificate]
\label{prop:qp-certificate}
For every $s\in\mathbb R^d$,
\begin{equation}
    D_q(s)\le \rho_q^{\mathrm{box}}
             \le \rho_q^{\mathrm{true}},
    \label{eq:qp-risk-order}
\end{equation}
where
\begin{equation}
    \rho_q^{\mathrm{true}}=\frac1d\min_{g\in\operatorname{conv}(A\cup\{b\})}\|y-g\|_2^2
    \label{eq:true-augmented-risk}
\end{equation}
is augmented coordinate-average risk for the fixed, complete candidate. Consequently,
\begin{equation}
    \Delta_{b,n}^\star
    \le R_n(f_{A,n}^\star)-\frac dn D_q(s).
    \label{eq:qp-dual-gain-upper}
\end{equation}
The first inequality remains valid for any finite value of $s$; optimizer
convergence is not required.
\end{proposition}

\begin{proof}
For every prediction vector $g$ and every $s$,
\begin{equation}
    \frac1d\|y-g\|_2^2
    \ge 2s^\top(y-g)-d\|s\|_2^2.
\end{equation}
Minimizing the left side over the convex hull of the incumbents and
$\mathcal B_q$, then maximizing $s^\top g$ on the right, gives
$D_q(s)\le\rho_q^{\mathrm{box}}$. Fixing the completion to the actual
candidate can only increase the optimum, which gives the second inequality.
Converting the lower bound to empirical-risk units and subtracting it from
incumbent empirical risk proves
\eqref{eq:qp-dual-gain-upper}.
\end{proof}

\subsection{Multiclass completion}
\label{app:multiclass-completion}

For Brier loss, let $b_{ik}=(p_{b,i})_k$ denote the candidate's
predicted probability for class $k$ of example $i$, with
$k=1,\ldots,K_i$. Concatenating the probability vectors gives
$d=\sum_{i=1}^n K_i$ entries, indexed by pairs $(i,k)$.
Querying example $i$ reveals all its $K_i$ probabilities.

The completion set $\mathcal B_q$ fixes $v_{ik}=b_{ik}$ for
$i\in\mathcal I_q$ and allows $L_{ik}\le v_{ik}\le U_{ik}$ for
$i\notin\mathcal I_q$, for every valid class $k$.
The corresponding QP constraints become
\begin{equation}
    \begin{aligned}
        t_{ik} &= \beta b_{ik}
            && (i\in\mathcal I_q),\\
        \beta L_{ik} &\le t_{ik}\le \beta U_{ik}
            && (i\notin\mathcal I_q),
    \end{aligned}
    \label{eq:multiclass-qp-constraints}
\end{equation}
for every $k=1,\ldots,K_i$.
Here, $y$ concatenates the one-hot targets, and each column of
$\mathbf A$ concatenates an incumbent's probability vectors.

The factor $d/n$ in \eqref{eq:qp-risk-scale} converts
coordinate-average loss to Brier loss per question, giving
the exact compatible gain
\begin{equation}
    R_n(f_{A,n}^\star)
    - \frac{d}{n}\rho_q^{\mathrm{box}}.
\end{equation}

Enforcing the known probability simplex would add only
the linear constraints
\begin{equation}
    t_{ik}\ge 0,
    \qquad
    \sum_{k=1}^{K_i} t_{ik}=\beta,
\end{equation}
preserving exactness for that set.

The experiments use the larger coordinate box $[0,1]^{K_i}$,
with $L_{ik}=0$ and $U_{ik}=1$. It contains the true candidate
and also allows extra completions that can delay rejection.
Completeness for this box therefore does not imply completeness
for all the available simplex information.

\section{Experimental setup and reproducibility}
\label{app:experimental-setup}

\subsection{Regression panel}
\label{app:regression-setup}

We use the 35 regression tasks in OpenML suite 353 (OpenML-CTR23)
\citep{fischer2023openmlctr23}. Each task is capped at 12,000 rows and split
into 60\% model training, 25\% screening data, and 15\% held-out data. Targets are
standardized from the training split, predictions are clipped to $[-4,4]$,
and screening set sizes range from 129 to 3,000 rows.

For each task and ten model seeds, we train three variants from ridge
regression, random forest, histogram gradient boosting, and a multilayer
perceptron. A case holds out one family: its third, highest-capacity recipe is
the candidate, while the incumbents are the first two variants from each of
the other three families. The panel therefore contains
\begin{equation}
    35\text{ tasks}\times10\text{ seeds}\times4\text{ families}
    =1{,}400\text{ candidate cases}.
    \label{eq:case-count}
\end{equation}

The primary target is the decision from the full screening set in Equation~\eqref{eq:screening-threshold}.
For the supplementary \emph{held-out useful} label, we fit simplex weights on the
full screening set for $A$ and $A+b$, freeze both fits, and evaluate their held-out
mean squared errors. For positive incumbent held-out risk, a candidate is useful when
\begin{equation}
    \frac{R_{A,\mathrm{test}}-R_{A+b,\mathrm{test}}}
         {R_{A,\mathrm{test}}}
    \ge0.01.
    \label{eq:heldout-label}
\end{equation}
There are 638 useful cases, or 45.57\%. Policies never receive held-out
labels, predictions, or risks.

The nominal budgets are
\begin{equation}
    B\in\{8,16,32,64,128,256,512,1024,2048\}.
\end{equation}
Budgets through 128 fit every screening set. A larger budget is truncated when the
screening set is smaller; the budget-trajectory figures use nominal caps on the horizontal axis.

\subsection{LLM panels}
\label{app:llm-setup}

The original panel contains the first eight models in
Table~\ref{tab:llm-roster}. In the task-family experiment, each model becomes
the candidate while the other seven are incumbents. In the model-family
experiment, the original eight remain fixed as set $A$ and each of the four
new-family models is added separately. All evaluated artifacts are pinned Q4
GGUF files. New-candidate predictions, caches, and run outputs use a separate
namespace and cannot enter set $A$ or one another's fits.

For background on the model families and releases listed in
Table~\ref{tab:llm-roster}, see the reports and technical articles
for Qwen3 \citep{qwen2025qwen3},
Llama 3 \citep{grattafiori2024llama3},
Gemma 4 \citep{gemmateam2026gemma4},
Ministral 3 \citep{liu2026ministral3},
Phi-4 Mini \citep{microsoft2025phi4mini},
Granite 4.1 \citep{ibm2026granite41},
OLMo 3 \citep{olmo2025olmo3},
Falcon-H1 \citep{zuo2025falconh1},
SmolLM3 \citep{bakouch2025smollm3},
MiniCPM4 \citep{minicpm2025minicpm4},
GLM-4 \citep{glm2024chatglm}, and
Apertus \citep{apertus2025apertus}.

The 15 task families are RACE-High, BLiMP adjunct-island, ToxiGen, WiC, PAWS-X
English, XNLI English, WMDP-Cyber, XCOPA Italian, TruthfulQA MC1, QASC,
QuaRTz, QuAIL, AG News, IMDb, and ReClor
\citep{lai2017race,warstadt2020blimp,hartvigsen2022toxigen,
pilehvar2019wic,yang2019pawsx,conneau2018xnli,li2024wmdp,
ponti2020xcopa,lin2022truthfulqa,khot2020qasc,tafjord2019quartz,
rogers2020quail,zhang2015character,maas2011sentiment,yu2020reclor}.
Passages, premises, and reused supporting facts remain intact across the
screening set and held-out split. Every task supplies a screening set of 192 rows. The
primary held-out set has 320 rows for 12 tasks, 318 for XNLI, and 192 for
XCOPA and ReClor.

A constrained next-token query returns a normalized probability vector over
the rendered choices. A tokenizer and template census checked that every
request fits the 4,096-token context. We fit incumbent and augmented convex
weights on the 192-row screening set and threshold the resulting gain for the primary
decision. For the supplementary held-out usefulness label, we freeze those
weights and require at least a one-percent reduction in held-out Brier loss.
No held-out candidate probability enters a query path, fallback fit, or
certificate.

The task-family gate contains $15\times8=120$ cases; the model-family gate
contains $15\times4=60$. Each case is replayed on ten deterministic nested
paths at caps $16,32,64,128,$ and $192$. The resulting 1,200 and 600 path
decisions are repeated measurements rather than independent cases.

\begin{table}[!ht]
    \centering
    \caption{Supplementary model metrics for the introduction's
    accuracy-oracle comparison in Table~\ref{tab:llm-roster}. Standalone Brier
    is the task-macro mean summed multiclass loss per question. Selected-task
    counts use strictly positive held-out accuracy gain; these counts determine
    the oracle call savings in Table~\ref{tab:llm-roster}.}
    \label{tab:llm-roster-details}
    \setlength{\tabcolsep}{6pt}
    \begin{tabular}{@{}llrr@{}}
        \toprule
        Model & Family & Standalone Brier $\downarrow$ & \shortstack{Accuracy-oracle\\selected tasks / 15} \\
        \midrule
        Qwen3 Instruct 2507 & Qwen & 0.4091 & 7 \\
        Llama 3.2 Instruct & Llama & 0.4976 & 2 \\
        Gemma 4 E2B IT & Gemma & 0.4804 & 4 \\
        Ministral 3 Instruct 2512 & Mistral & 0.4081 & 4 \\
        Phi-4 Mini Instruct & Phi & 0.4390 & 2 \\
        Granite 4.1 & Granite & 0.5074 & 5 \\
        OLMo 3 Instruct & OLMo & 0.4801 & 5 \\
        Falcon-H1 Instruct & Falcon & 0.3391 & 9 \\
        \midrule
        SmolLM3 & SmolLM & 0.5031 & 4 \\
        MiniCPM 4.1 & MiniCPM & 0.4379 & 5 \\
        GLM-4 0414 & GLM & 0.3349 & 8 \\
        Apertus Instruct 2509 & Apertus & 0.5862 & 1 \\
        \bottomrule
    \end{tabular}
\end{table}

The original-eight aggregate reaches 81.11\% held-out accuracy, compared with
78.59\% for single-model selection on the screening set, and lowers macro Brier from
0.3259 to 0.2662. The single-model baseline chooses one original model per task
by Brier loss on the screening set. The aggregate fits task-specific convex weights on the
screening set before held-out evaluation.

Table~\ref{tab:llm-roster} reports task-macro standalone accuracy and the
numbers of tasks helped, hurt, or unchanged by adding each candidate.
Effects compare held-out top-1 accuracy with and without the candidate,
after fitting both sets of convex weights by Brier loss on the screening set.
The parenthesized changes average only over the tasks in their respective
helped or hurt group; unchanged means equal accuracy, not identical
probabilities. The three counts sum to 15. Table~\ref{tab:llm-roster-details}
gives full model names, supporting Brier scores, and oracle task counts.
The accuracy oracle queries the candidate on all 192 screening rows of each
helped task and skips hurt and unchanged tasks. Its call savings are
\begin{equation}
    100\,\frac{15-n_{\mathrm{helped}}}{15},
    \label{eq:oracle-call-savings}
\end{equation}
with incumbent predictions already cached. This
assumes knowledge of held-out outcomes before querying and measures neither
an implementable screening policy nor total deployment savings.

Table~\ref{tab:llm-brier-selective-addition} gives the parallel analysis for
the supplementary held-out Brier usefulness label. An original model is compared with
the other seven, whereas a new-family candidate is compared with the fixed
original eight. No joint 12-model ensemble is fitted. Names, family labels,
and nominal size classes follow the frozen repository metadata; size is not
GGUF file size. E2B retains Google's effective-parameter size designation
for Gemma 4.

\begin{table}[!htbp]
    \centering
    \caption{Descriptive effects under the supplementary held-out Brier
    usefulness label. A task is Brier-useful when the
    augmented aggregate fitted on the full screening set reduces held-out Brier loss by at least
    one percent relative to the incumbent. Entries are task-macro means of
    the signed relative Brier reduction; positive is better. ``Useful only''
    assigns zero change to other tasks, so both policies retain the same
    15-task denominator. Within the two reduction columns, bold marks the
    larger result in each row. Always adding can win when excluded tasks have
    a positive net contribution, including gains below the one-percent
    threshold. This is a held-out-oracle diagnostic, not an operational
    routing policy.}
    \label{tab:llm-brier-selective-addition}
    \setlength{\tabcolsep}{6pt}
    \begin{tabular}{@{}lrrr@{}}
        \toprule
        Model & \shortstack{Brier-useful\\tasks / 15}
        & \shortstack{Relative Brier\\reduction (\%)\\always add}
        & \shortstack{Relative Brier\\reduction (\%)\\useful only} \\
        \midrule
        \multicolumn{4}{@{}l}{\textit{Original panel}} \\
        Qwen3 Instruct 2507 & 8 & +1.959 & \textbf{+2.554} \\
        Llama 3.2 Instruct & 3 & $-0.052$ & \textbf{+0.308} \\
        Gemma 4 E2B IT & 2 & +0.005 & \textbf{+0.218} \\
        Ministral 3 Instruct 2512 & 4 & \textbf{+1.261} & +1.128 \\
        Phi-4 Mini Instruct & 1 & +0.159 & \textbf{+0.245} \\
        Granite 4.1 & 3 & +0.579 & \textbf{+0.780} \\
        OLMo 3 Instruct & 3 & +0.963 & \textbf{+1.334} \\
        Falcon-H1 Instruct & 12 & +9.570 & \textbf{+10.044} \\
        \midrule
        \multicolumn{4}{@{}l}{\textit{New-family candidates (held out from development)}} \\
        SmolLM3 & 0 & $-0.013$ & \textbf{+0.000} \\
        MiniCPM 4.1 & 5 & +1.576 & \textbf{+1.581} \\
        GLM-4 0414 & 11 & \textbf{+7.604} & +7.555 \\
        Apertus Instruct 2509 & 4 & +0.658 & \textbf{+0.728} \\
        \midrule
        \textbf{Across-candidate mean} & --- & +2.022 & \textbf{+2.206} \\
        \bottomrule
    \end{tabular}
\end{table}

\subsection{Matched protocols and uncertainty}
\label{app:uncertainty}

The fixed-budget sampling comparison has no early stopping. Uniform sampling
uses a nested uniform prefix. The two residual-aware methods form four strata
from incumbent loss or the alignment envelope, apply the same stabilized
allocation, and fit the same post-stratified full-simplex objective. Every
method reaches the cap and predicts useful when fitted relative gain is at
least one percent.

The stopping comparison fixes the alignment-stratified path, checkpoints,
post-stratified fit, and one-percent fallback. No stopping always reaches the
cap. Two-negative stability rejects after two consecutive distinct
positive-query checkpoints predict useless and never accepts early. The
alignment-completion rule uses Corollary~\ref{cor:alignment-completion}; the
coordinate-box QP uses Proposition~\ref{prop:qp-certificate}. Unresolved
cases reach the shared cap and fallback.

The conditional-scenario comparison keeps that path and fallback fixed. It is
inactive through 64 queries and considers
$\delta\in\{0.01,0.03,0.05,0.10,0.20\}$. Each active checkpoint requests up
to 512 complete posterior draws. A rejected checkpoint evaluates all 512;
the LLM replay can short-circuit a non-rejecting checkpoint after exceeding the
active material-count cutoff. The regression replay reuses random numbers
across checkpoints and thresholds on a path. The LLM replay uses a
checkpoint-specific seed and reuses draws only across thresholds within that
checkpoint. A scenario is material when its exactly reoptimized
relative gain is at least one percent. Rejection requires the pointwise
one-sided 99\% Clopper-Pearson upper bound \citep{clopper1934use} on the material-scenario probability
to be below $\delta$. The plotted points are paired replay means on the same
ten paths per case; $\delta$ was swept rather than chosen on an independent
validation panel.

For stopping rule $s$ at cap $B$, the primary metrics are $100A_{s,B}$,
where $A_{s,B}$ is the fraction of decisions matching reaggregation on the full screening set, and call saving relative to the full screening set
\begin{equation}
    S^{\mathrm{full}}_{s,B}
    =100\left(1-\frac{Q_{s,B}}{\bar n}\right).
    \label{eq:full-screening-saving}
\end{equation}
Here $Q_{s,B}$ is mean calls and $\bar n$ is mean screening set size under the
panel weighting below. Regression has $\bar n=1{,}712.23$ rows, while both
LLM panels have $\bar n=192$. For no stopping,
$Q_{\mathrm{no\ stop},B}$ is the panel mean of $\min(B,n)$.
The ratio of means counts the fraction of total candidate calls avoided
under that weighting; it is not the mean of per-case saving fractions.

Supplementary stopping controls and the posterior threshold sweep instead
isolate early stopping at a fixed cap. They report matched-cap call saving
\begin{equation}
    100\frac{Q_{\mathrm{no\ stop},B}-Q_{s,B}}
             {Q_{\mathrm{no\ stop},B}},
    \label{eq:relative-saving}
\end{equation}
and change in screening accuracy
\begin{equation}
    100\left(A_{s,B}-A_{\mathrm{no\ stop},B}\right),
    \label{eq:screening-agreement-change}
\end{equation}
These supplementary quantities use capped no stopping as their baseline;
they are distinct from the absolute metrics in Figures
\ref{fig:stratification-and-stopping-by-budget} (a) and~\ref{fig:stratification-and-stopping-by-budget} (b).

All panels use ten nested paths per case and 5,000 paired hierarchical
bootstrap draws. Regression estimates average paths within cases, then seeds
and candidate families within datasets, and finally weight the 35 datasets
equally. Its bootstrap resamples datasets, seeds within datasets, and paths
within seeds while retaining the four candidates jointly. The LLM bootstrap
resamples task families and paths within family while retaining the fixed
candidate panel. Draws are shared across methods and axes, and ratios are
computed inside each draw. The independent LLM task dimension therefore has
15 clusters. The four new candidates are a fixed purposive set, so their
intervals do not support a population-level model-family claim.

An always-reject rule gives zero-query held-out usefulness-label accuracies
of 54.43\% in regression,
70.00\% for unseen LLM tasks, and 66.67\% for unseen LLM models. It is omitted
from the stopping comparison because it is a degenerate prevalence baseline
that never screens a candidate. Reaggregation on the full screening set is a
maximum-information reference, not perfect held-out ground truth. The
regression sampling and stopping controls are exploratory replays on an already
studied panel.

\subsection{Certificate implementation}
\label{app:qp-implementation}

\paragraph{Conservative numerical evaluation.}

To certify rejection numerically, we need a lower bound on the
optimistic risk. A solver's objective value alone does not provide
that bound. The implementation evaluates an explicit weak-dual
lower bound $D_q(s)$ at finite dual-feasible witness vectors $s$.
It satisfies
\begin{equation}
    D_q(s)\le \rho_q^{\mathrm{box}}
\end{equation}
in exact arithmetic, even without solver convergence
(Proposition~\ref{prop:qp-certificate}).

We subtract a small numerical cushion and retain the largest
accepted lower bound along the path, denoted
$\underline{\rho}_q^{\mathrm{QP}}$.
To express this bound as loss per screening example, we use
\begin{equation}
    \underline{R}_q^{\mathrm{QP}}
    = \frac{d}{n}\,\underline{\rho}_q^{\mathrm{QP}}.
    \label{eq:qp-risk-scale}
\end{equation}
Here, $n$ counts screening examples and $d$ counts prediction
coordinates; for scalar regression, $d=n$.
Underlines and overlines denote lower and upper bounds.

Subtracting this lower bound from the incumbent loss gives an
upper bound on the improvement achievable by adding the candidate:
\begin{equation}
    \overline{\Delta}_q^{\mathrm{QP}}
    = \max\!\left\{
        0,\;
        R_n(f_{A,n}^\star)
        - \underline{R}_q^{\mathrm{QP}}
      \right\}.
    \label{eq:qp-gain-upper}
\end{equation}
The final policy is to reject when
$\overline{\Delta}_q^{\mathrm{QP}}<\tau_n$.
A solver failure returns no new bound.

Because approximate witnesses may give weaker bounds, the
implementation can reject later than the exact optimum and does
not inherit its completeness. The floating-point cushion guards
against numerical error but is not a formal rounding-error proof.
The implementation details below specify the tolerances and
failure handling.

For Brier loss, each query reveals a full probability vector;
the LLM implementation bounds unobserved probabilities individually
in $[0,1]$, preserving safety but potentially delaying rejection
(Appendix~\ref{app:multiclass-completion}).

\paragraph{Solver and dual witnesses.}
The implementation eliminates $t$ entry by entry. In scalar regression, for an
unqueried example it clips $y_i-(A\alpha)_i$ to $[\beta L_i,\beta U_i]$.
For multiclass predictions, the same clipping applies independently to each
class entry $(i,k)$ of an unqueried example, using $[\beta L_{ik},\beta U_{ik}]$;
queried examples have all entries fixed at $t_{ik}=\beta b_{ik}$.
Thus SLSQP optimizes only the $m+1$ simplex weights with an analytic gradient.
From both the starting point
and the projected solver output, it forms a residual, divides it by $d$, and
uses the result as a witness in \eqref{eq:qp-weak-dual}. If these witnesses
form the set $\mathcal S_q$ at checkpoint $q$, the returned values are
\begin{equation}
    \ell_q
    =\max\left\{0,\ \max_{s\in\mathcal S_q}D_q(s)
                         -\varepsilon_{\mathrm{num}}\right\},
    \qquad
    \underline\rho_q^{\mathrm{QP}}
    =\max_{q'\le q}\ell_{q'}.
    \label{eq:qp-checkpoint-lower}
\end{equation}
The second maximum carries lower-bound values from earlier checkpoints; it does
not reevaluate their witnesses under the new completion box. This remains
valid because querying an example fixes its entire prediction vector,
shrinks $\mathcal B_q$, and makes
$\rho_q^{\mathrm{box}}$ nondecreasing.

\paragraph{Numerical tolerances and failure handling.}
We use $\varepsilon_{\mathrm{num}}=10^{-12}+10^{-10}S$, where $S$ is the largest of
one, the mean squared outcomes, the mean squared incumbent predictions, and the
squared prediction bounds. The subtraction is a floating-point guard, not an
interval-arithmetic proof. SLSQP uses a $10^{-12}$ objective tolerance and at
most 300 iterations. A solver exception returns no new lower bound. Nonfinite
or numerically inconsistent witness values are discarded, while invalid inputs
stop the run. A nonconverged projected iterate may still yield a valid weak
dual value because Proposition~\ref{prop:qp-certificate} holds for every finite
witness. A later solver failure leaves any earlier valid path-level bound in
place.

\paragraph{Evaluation schedule and computational cost.}
The policy checks the certificate at zero queries, every 32 queries, and each
budget before the census. The census uses exact reaggregation on the screening set.
Eliminating $t$ leaves $m+1$ optimization variables. One objective, gradient,
or dual evaluation costs $O(dm)$. With at most $I=300$ SLSQP iterations, a
conservative per-checkpoint bound is $O(I(dm+m^3))$ time and
$O(dm+m^2)$ storage. In the Brier experiments, $d$ is the number of valid
class coordinates rather than the number of questions.

\subsection{Stopping-policy overhead and inference-time reference}
\label{app:runtime-break-even}

Candidate-call savings translate into wall-clock savings only when the avoided
inference costs more than the additional stopping-policy work. For rule $s$,
we summarize this threshold by the break-even candidate time
\begin{equation}
    t_{\mathrm{BE}}(s)
    =\frac{T_s}
           {\sum_p\left(Q_{\mathrm{no\ stop},p}-Q_{s,p}\right)},
    \label{eq:runtime-break-even}
\end{equation}
where $T_s$ is the aggregate rule-specific runtime, measured or estimated as
described below, over the evaluated paths. The denominator counts calls
avoided relative to matched capped no stopping, isolating the additional
savings due to the stopping rule.
If marginal candidate inference exceeds $t_{\mathrm{BE}}(s)$, its estimated time
saved exceeds the estimated policy overhead. Paths with no saving remain in the
runtime numerator, so the calculation includes unsuccessful stopping checks.

The manuscript record places both original timing sources on a MacBook Pro with
an Apple M5 Pro processor (18 CPU cores) and 64 GB of unified memory. It
reports Python 3.12, NumPy 2.5.2, SciPy 1.18.1, and one-thread BLAS/OpenMP for
the safe-QP benchmark. No compact raw safe-QP timing manifest or original
inference logs were retained. Those timing values and protocol metadata are
therefore historical manuscript provenance and cannot be independently
reproduced from the repository; only the posterior-overlay timings below are
now retained compactly. The original record reports five warmed repetitions
over the exact frozen LLM paths and the median total time. The timed work includes coordinate
construction, the zero-query certificate, warm starts, and every SLSQP
certificate solve reached before stopping. It excludes alignment-path
construction and the final post-stratified refit, which are shared with the
matched no-stop comparison. Every reproduced QP stop matched the frozen path
record.

\begin{table}[!ht]
    \centering
    \caption{Historical QP overhead and panel-level break-even inference time at
    the 192-call cap. Times are totals over all paths, not per-path times.}
    \label{tab:qp-break-even-panel}
    \setlength{\tabcolsep}{6.7pt}
    \begin{tabular}{@{}lrrrr@{}}
        \toprule
        Panel & Paths & Calls avoided & QP time, median [range] (s)
              & $t_{\mathrm{BE}}$ (ms/call) \\
        \midrule
        Unseen LLM tasks  & 1{,}200 & 53{,}056 & $4.590\ [4.574,4.598]$ & 0.087 \\
        Unseen LLM models &   600 & 24{,}672 & $2.383\ [2.379,2.385]$ & 0.097 \\
        \bottomrule
    \end{tabular}
\end{table}

Collection notes place the posterior timing audit on the same hardware but in
the current locked environment: Python 3.12.13, NumPy 2.5.2, SciPy 1.18.0,
pandas 2.3.3, and
Matplotlib 3.11.1. These versions were inspected at evidence collection rather
than embedded in the temporary run manifests, and their BLAS/OpenMP limits were
not recorded. We timed the $\delta=0.05$ overlay for Qwen3 4B on its 15 cases
and ten paths per case, using one process worker, 512 scenarios, one warm-up,
and five measured repetitions.
The marginal-box overlay takes a median 1.0881 s
$[1.0778,1.0957]$ and performs 600 posterior-bound QP solves. The conditional
overlay takes 11.3183 s $[11.0882,11.6756]$ and performs 30,067 exact scenario
refits per repetition. These overlays read the stored safe-QP trace. Adding
the approximately 0.6406 s implied by the rounded 0.286 ms/call safe-QP value
and 2,240 avoided calls gives the implied integrated estimates in
Table~\ref{tab:policy-runtime-qwen}; shared work was not timed as one process.

Table~\ref{tab:policy-runtime-qwen} summarizes the Qwen3 4B comparison. A safe
checkpoint solves one optimistic QP. A marginal posterior box adds one
posterior-bound QP per threshold and checkpoint. Conditional scenarios require
an exact augmented-ensemble refit per retained draw, with early abort when
rejection becomes impossible. The marginal box changes no Qwen stopping
outcome, whereas scenarios avoid 3,840 calls, 71.4\% more than the safe QP.
Their approximate 12.0 s policy total is about 80 ms per path and 3.7\% of the
estimated avoided inference time. Inference depends on hardware, batching,
model size, quantization, prompt and output length, and inference-time
reasoning; this calculation is an approximate implementation estimate.

\begin{table}[!ht]
    \centering
    \caption{Candidate-specific stopping-policy overhead for Qwen3 4B at the
    192-call cap. Break-even time divides policy computation by calls avoided;
    the last column divides the 83.45 ms mean inference request by that value.
    Posterior totals combine replay time with the safe-QP time
    implied by its previously measured break-even value and are rounded because shared work was not timed as
    one integrated implementation. Replay timing includes input loading and
    validation but excludes artifact writing and plotting. These are overhead
    margins, not end-to-end speedups.}
    \label{tab:policy-runtime-qwen}
    \setlength{\tabcolsep}{5pt}
    \begin{tabular}{@{}lrrrr@{}}
        \toprule
        Policy & \shortstack{Calls\\avoided} & \shortstack{Est. policy\\time (s)}
               & \shortstack{$t_{\mathrm{BE}}$\\(ms/call)} & \shortstack{Inference /\\overhead} \\
        \midrule
        Safe optimistic QP & 2,240 & $\approx0.641$ & 0.286 & $292\times$ \\
        \shortstack[l]{Marginal posterior box,\\$\delta=.05$} & 2,240 & $\approx1.7$
            & $\approx0.77$ & $\approx108\times$ \\
        \shortstack[l]{Conditional scenarios,\\$\delta=.05$} & 3,840 & $\approx12.0$
            & $\approx3.11$ & $\approx26.8\times$ \\
        \bottomrule
    \end{tabular}
\end{table}

For batch-throughput context only, the six-worker, all-$\delta$ replays take
263.6 s for 14,000 regression paths and 54.8 s for 1,800 LLM paths. Those runs
amortize setup across five thresholds and use early abort, so they are not
single-policy online latency measurements.

For the inference reference, the frozen runs used \texttt{llama.cpp} 0.2.0,
build 10566 (commit \texttt{bb4caa754}), six inference threads, one slot, and a
4,096-token context. Each request produced a constrained one-token score.
We read \texttt{slot print\_timing: total time} from the server logs; this
excludes model loading, client work, and network latency. After removing one
warm-up request from each original-panel log and 11 diagnostic or preflight
requests from each new-model log, every model contributes 7,422 production
requests. The logs preserve the original sequential request order and its
KV/prefix-cache behavior.

Table~\ref{tab:roster-runtime} pairs these request times with a break-even value
computed by restricting Equation~\ref{eq:runtime-break-even} to each candidate.
The ratio uses the corpus-wide mean request time as the per-call reference; it
does not claim that the evaluation is that many times faster end to end.

\begin{table}[!ht]
    \centering
    \caption{Historical frozen-run one-token inference timing and candidate-specific QP
    break-even time on the same machine. The Qwen3 row gives the smallest
    mean-to-break-even ratio in the roster.}
    \label{tab:roster-runtime}
    \setlength{\tabcolsep}{7pt}
    \begin{tabular}{@{}lrrrr@{}}
        \toprule
        Model & \shortstack{Median\\request (ms)} & \shortstack{Mean\\request (ms)}
              & \shortstack{Break-even\\(ms)} & \shortstack{Mean /\\break-even} \\
        \midrule
        \multicolumn{5}{@{}l}{\textit{Original panel}} \\
        Qwen3 4B          & 57.94  & 83.45     & 0.286 & $292\times$ \\
        Llama 3.2 3B      & 44.22  & 62.17     & 0.064 & $971\times$ \\
        Gemma 4 E2B       & 63.28  & 87.83     & 0.057 & $1{,}533\times$ \\
        Ministral 3 3B    & 47.93  & 70.27     & 0.102 & $689\times$ \\
        Phi-4 Mini 3.8B   & 52.01  & 72.73     & 0.055 & $1{,}315\times$ \\
        Granite 4.1 3B    & 54.14  & 74.12     & 0.063 & $1{,}179\times$ \\
        OLMo 3 7B         & 136.96 & 218.39    & 0.074 & $2{,}957\times$ \\
        Falcon-H1 7B      & 203.62 & 1{,}300.85 & 0.505 & $2{,}577\times$ \\
        \addlinespace
        \multicolumn{5}{@{}l}{\textit{New-family candidates}} \\
        SmolLM3 3B        & 48.18  & 64.12  & 0.066 & $966\times$ \\
        MiniCPM 4.1 8B    & 106.64 & 165.72 & 0.092 & $1{,}798\times$ \\
        GLM-4 9B          & 114.05 & 166.00 & 0.209 & $796\times$ \\
        Apertus 8B        & 101.84 & 158.01 & 0.091 & $1{,}744\times$ \\
        \bottomrule
    \end{tabular}
\end{table}

Even this closest case is 292 times above break-even, corresponding to QP
overhead of about 0.34\% of the estimated avoided Qwen3 inference time. Models
with slower marginal inference would have a larger margin if their stopping
trajectory were held fixed. Absolute timings and the ratio are machine-,
batching-, request-order-, and implementation-dependent, whereas candidate
calls remain the more portable primary measure. The current requests also
generate only one constrained token; longer outputs or inference-time
reasoning would change the inference side of the comparison, not the QP
program itself.

\subsection{Provenance and sensitivity checks}
\label{app:provenance}

The supplementary manifests bind the configurations, task and model sources,
repository revisions, prediction shards, output filenames, and SHA-256 hashes.
The loss sampler and matched stopping comparison are post-hoc cache replays;
they do not run new inference. The exact variance audit is also cache-only and
adds no refit, stopping rule, or outcome bootstrap. The model-family cache and run namespace is
separate from the original panel, and the four new candidates are evaluated
one at a time.

Before the model-family run, a compatibility preflight scored WiC[0] for
SmolLM3 and Apertus and issued diagnostic score requests for MiniCPM. GLM was
tested only on synthetic input. The preflight computed no usefulness, QP,
aggregate, or method result and changed none of the roster, tasks, methods,
budgets, threshold, or success rules.

Thirteen of 60 model-panel held-out usefulness labels change on the
smaller 63/64-row held-out sensitivity set. Excluding QASC leaves the LLM
headline conclusions unchanged. Benchmark contamination in model pretraining
cannot be ruled out, and the model-family test reuses the task surface from the
task-family experiment. These checks limit the claim to the fixed artifacts,
tasks, and reductions in queries to the screening set reported here.

\section{Sampling and stopping controls}
\label{app:component-ablations}

The supplementary accuracy results in this appendix measure agreement with
the held-out usefulness label. Matched-cap savings isolate early stopping
using Equation~\eqref{eq:relative-saving}; the main figures use the full
screening set as their reference.

\subsection{QP and combined stopping}

On the frozen paths, combining the QP and alignment-completion bounds gives
the same calls and decisions as QP-only at every reported budget. At the
largest cap, the QP rejects earlier or uniquely while saving at least one call
on 4,511 regression paths, 512 unseen-task paths, and 237 unseen-model paths;
alignment does so on none. This is an empirical ordering of the implemented
bounds. Equation~\eqref{eq:qp-dominates-alignment} orders exact bounds when
they use the same completion set.

\subsection{Incumbent-loss and alignment query paths}

The main sampling comparison disables stopping. A separate replay asks
whether its conclusion changes after both schedules use the same QP-only
certificate. Table~\ref{tab:qp-path-control} reports the largest common cap;
positive call differences mean that the incumbent-loss path uses more calls.

\begin{table}[!htbp]
    \centering
    \caption{Post-hoc query-schedule comparison with the coordinate-box QP
    held fixed. Accuracy scores held-out usefulness labels. Values are
    paired point differences with 95\% intervals. No
    accuracy contrast is resolved. The alignment schedule uses modestly fewer
    calls in regression; the LLM call differences are unresolved.}
    \label{tab:qp-path-control}
    \setlength{\tabcolsep}{4pt}
    \begin{tabular}{@{}lrr@{}}
        \toprule
        Panel and cap & \shortstack{Loss $-$ alignment\\held-out accuracy (pp)}
                      & Loss $-$ alignment calls \\
        \midrule
        Regression, 2048 & $+0.01\ [-0.05,0.09]$
                         & $+5.61\ [2.44,9.35]$ \\
        LLMs: unseen tasks, 192 & $0.00\ [0.00,0.00]$
                                & $-1.20\ [-2.91,0.24]$ \\
        LLMs: unseen models, 192 & $0.00\ [0.00,0.00]$
                                 & $-0.11\ [-2.03,1.55]$ \\
        \bottomrule
    \end{tabular}
\end{table}

This check supports using the frozen alignment schedule for the matched
stopping comparison, but not a claim that it is universally better. The two
schedules have similar accuracy, and their relationship is already explained
by Proposition~\ref{prop:loss-alignment-headroom}.

\subsection{Residual-aware assignment placebo}

A second sampling check asks whether residual-aware row assignment matters
relative to a matched permutation. It deterministically reassigns complete
alignment-bound pairs while preserving their multiset, the four stratum sizes,
the stabilized allocation, nested budgets, and the post-stratified convex
refit. No method in this ablation stops early.

\begin{table}[!htbp]
    \centering
    \caption{Post-hoc row-assignment placebo at 64 calls. Accuracy scores
    held-out usefulness labels and is reported in percent; effects are percentage-point differences. The contrast isolates
    the row-to-bound assignment
    inside the fixed four-stratum, weighted-refit pipeline. It does not compare
    alignment with every possible row-difficulty score.}
    \label{tab:alignment-placebo}
    \setlength{\tabcolsep}{3.5pt}
    \begin{tabular}{@{}lrrrr@{}}
        \toprule
        Panel & Uniform & Placebo & \shortstack{Alignment-\\stratified}
              & \shortstack{Assignment effect\\{[95\% interval]}} \\
        \midrule
        Regression & 79.22 & 77.69 & 81.88 & $+4.19\ [2.75,5.63]$ \\
        LLMs: unseen tasks & 76.83 & 77.33 & 81.50
            & $+4.17\ [-0.92,9.33]$ \\
        LLMs: unseen models & 78.17 & 77.67 & 81.00
            & $+3.33\ [-2.34,8.83]$ \\
        \bottomrule
    \end{tabular}
\end{table}

The assignment effect has the same sign in all three panels, but only the
regression interval excludes zero. Placebo minus uniform is $-1.53$ points in
regression, $+0.50$ on unseen task families, and $-0.50$ on unseen model
families; all but the regression contrast have intervals that include zero.
Unlike the incumbent-loss comparison, this placebo does not distinguish
alignment geometry from a natural incumbent-difficulty score. It reuses the
frozen panels and is not an independent confirmation.

\subsection{Other screening controls}

The LLM-only scalar mixture freezes the incumbent aggregate fitted on the full screening set and
fits one mixing weight between it and the candidate on a uniform prefix. At
64 calls it reaches 75.67\% held-out usefulness-label accuracy on unseen task families and 74.00\% on
unseen model families, compared with 76.83\% and 78.17\% for uniform
full-simplex refitting. The model-gate difference is $-4.17$ points with
interval $[-7.00,-1.50]$. This is a useful simple baseline, but it is too
restrictive when the incumbent weights should also move.

A standalone optimistic QP queries rows in descending order of their known
alignment-interval width and continues until certified rejection or the exact
decision from the full screening set. On the regression panel it reproduces every
decision from the full screening set at 1,356 mean calls. This gives one naturally
stopping reference point rather than a budget curve.

\subsection{Supplementary held-out diagnostics}
\label{app:heldout-diagnostics}

The primary comparison of screening accuracy is complemented by the held-out
usefulness label defined in Equation~\eqref{eq:heldout-label}. In the original
LLM panel, every model is useful on at least one task family, and 29 of the
36 useful additions have worse standalone Brier loss than their incumbent
aggregate (Table~\ref{tab:llm-brier-selective-addition}).

With early stopping disabled, the two scores improve held-out decision accuracy
at 64 calls by
2.66--3.16 percentage points in regression and 3.92--4.67 points on unseen
LLM tasks; all four paired 95\% intervals exclude zero. The unseen-model
estimates are less precise: the loss-minus-uniform interval touches zero and
the alignment-minus-uniform interval crosses it. Direct loss--alignment
contrasts at 64 calls remain unresolved: $+0.50$ points $[-0.25,1.22]$ in
regression, $-0.75$ $[-3.08,1.42]$ on unseen tasks, and $+1.83$
$[-1.17,5.00]$ on unseen models.

At the largest caps, QP and capped no stopping have the same held-out
usefulness-label accuracy: 89.24\%, 85.00\%, and 83.33\% in regression,
unseen LLM tasks, and unseen LLM models. Relative to QP, two-negative
stability changes this accuracy by $-8.10$ points $[-11.22,-4.94]$,
$-2.75$ $[-6.17,0.83]$, and $-3.83$ $[-8.67,0.83]$, respectively.
The LLM intervals include zero. Posterior scenarios at $\delta=0.05$
change held-out accuracy by $-4.17$, $0.00$, and $-0.17$ points relative
to capped no stopping (Table~\ref{tab:posterior-scenario-full-delta}).

The decision from the full screening set itself disagrees with held-out usefulness in
150 of 1,400 regression cases. It misses 7 of 36 held-out-useful additions
on unseen LLM tasks and 5 of 20 on unseen models, giving useful-candidate
recall of 80.6\% and 75.0\%. It also accepts 11 of 84 held-out-negative
task-panel cases and 5 of 40 held-out-negative model-panel cases.
The certificate cannot repair this gap between full-evaluation decisions and held-out usefulness.
Thirteen of 60 model-panel labels change on the smaller held-out
sensitivity split; intervals for held-out accuracy condition on the observed
labels and exclude this label uncertainty.

For completeness, largest-cap QP savings relative to capped no stopping
are 5.89\%, 23.03\%, and 21.42\% overall, and 11.52\%, 34.54\%, and
32.13\% among paths for candidates rejected by full evaluation. Alignment completion saves
0.81\%, 12.90\%, and 11.17\% overall against the same capped baseline.
At the 192-row LLM census, per-candidate QP savings range from 4.44\%
(Falcon-H1) to 32.89\% (Phi-4 Mini) in the task panel, and from 10.22\%
(GLM-4) to 30.00\% (SmolLM3) in the model panel. These are descriptive
ranges without candidate-level uncertainty estimates.

\section{Posterior-scenario construction and diagnostics}
\label{app:posterior-scenarios}

\subsection{Scenario decision rule}

Let $\Pi_q(\,\cdot\mid\mathcal O_q)$ denote a working predictive distribution
over complete candidate outputs after the queried observations $\mathcal O_q$.
For a draw $Z$, let $\Delta^\star(Z)$ be its gain on the screening set after exact augmented
convex reaggregation. Assuming $R_n(f_{A,n}^\star)>0$, define
$G^\star(Z)=\Delta^\star(Z)/R_n(f_{A,n}^\star)$ and
\begin{equation}
    p_q=\Pr_{Z\sim\Pi_q}\{G^\star(Z)\ge\gamma\mid\mathcal O_q\},
    \qquad\gamma=0.01.
    \label{eq:posterior-material-probability}
\end{equation}
At a rejecting checkpoint, all $N_{\mathrm{MC}}=512$ scenarios are evaluated.
If $K_q$ attain the material gain, compute the one-sided 99\%
Clopper--Pearson upper bound \citep{clopper1934use} $U_{0.99}(K_q,N_{\mathrm{MC}})$ and reject when
\begin{equation}
    U_{0.99}(K_q,N_{\mathrm{MC}})<\delta.
    \label{eq:posterior-scenario-rejection}
\end{equation}
A checkpoint can declare non-rejection early once the material count makes
this inequality impossible. Failed refits count as material. The binomial
bound controls finite simulation error at one checkpoint under the working
distribution. It supplies neither coverage under model misspecification nor
a time-uniform guarantee for repeated inspections.

\subsection{Working predictive distributions}

The scenario rule uses a working predictive model for missing outputs; it leaves the frozen
alignment-stratified path, checkpoints, one-percent material-gain threshold,
and exact ensemble refit unchanged. It is inactive through 64 calls and can
therefore first change a reported decision at the next budget. For regression,
the same random numbers are reused across checkpoints and values of $\delta$ on
each path. The LLM replay uses checkpoint-specific draws and shares them only
across values of $\delta$ within that checkpoint.
Observed candidate outputs are copied unchanged into every completion.

For regression, write the candidate prediction as the incumbent aggregate plus
a candidate deviation. This deviation follows a Bayesian linear model with 11 known row
features: four frozen-stratum indicators, the standardized empirical residual, and six standardized incumbent deviations from their aggregate. The
coefficient prior has mean zero and precision $4I$ conditional on the noise
variance, while that variance has inverse-gamma parameters
$(\alpha_0,\beta_0)=(2,16/3)$. Each scenario shares one coefficient and
variance draw across the screening set, draws the remaining row noise conditionally
independently, inflates both posterior scales by 1.3, and clips the resulting
predictions to $[-4,4]$. Clipping approximates the original clipped prediction
pipeline; it is not a censored-data posterior.

For the LLM panels, candidate probabilities are compared with the known
incumbent aggregate in centered-log-ratio coordinates, using $10^{-4}$ additive
smoothing. Within each frozen stratum, separate scalar normal--inverse-gamma
models are updated for the target coordinate and the pooled non-target
coordinates. Their prior parameters are
$(\mu_0,\kappa_0,\alpha_0,\beta_0)=(0,4,2,4)$, with the predictive scale again
inflated by 1.3. A scenario shares its role--stratum location and variance
draws, adds conditionally independent coordinate noise, and applies a softmax
to return each row to the probability simplex. The model thus learns
conditional location and scale, but not an arbitrary deviation shape or
dependence structure. The separate scalar updates use a working
independence approximation for candidate deviations in centered-log-ratio coordinates, whose
coordinates sum to zero within a row. Softmax ensures valid probability
vectors; the resulting predictive distribution is not claimed to be
calibrated.

For $N_{\mathrm{MC}}=512$ draws, the one-sided 99\% Clopper--Pearson condition in
Equation~\eqref{eq:posterior-scenario-rejection} allows at most 0, 6, 14, 35,
and 81 material scenarios for
$\delta=0.01,0.03,0.05,0.10,$ and $0.20$, respectively. Any failed refit is
counted as material. The full replays had no solver failures: regression
evaluated 24,942,080 complete scenarios with a batched exact active-face solver
and checked 83,256 fits independently with SLSQP; the LLM replay evaluated
1,075,946 of 2,249,216 requested scenarios, with early abort accounting for
the difference.

\subsection{Budget trajectories and threshold sensitivity}

Figure~\ref{fig:stratification-and-stopping-by-budget} (b) compares screening accuracy and calls saved
relative to the full screening set across budgets at the
representative post-hoc threshold $\delta=0.05$. The rule follows the safe QP
through 64 calls by construction, so only the 128- and 192-call LLM budgets
can differ. Once active, scenario call savings lie between QP and stability.
The corresponding screening accuracy exceeds that of a pointwise randomized
mixture of those two rules matched to the scenario mean calls: the descriptive
difference ranges from 0.62 to 4.66 points in regression, 2.28 to 2.86 on
unseen tasks, and 2.47 to 3.17 on unseen models. The mixing weight varies with
budget, and these contrasts have no uncertainty intervals. They establish
neither calibration nor population-level dominance.

\begin{figure}[H]
    \centering
    \includegraphics[width=\textwidth]{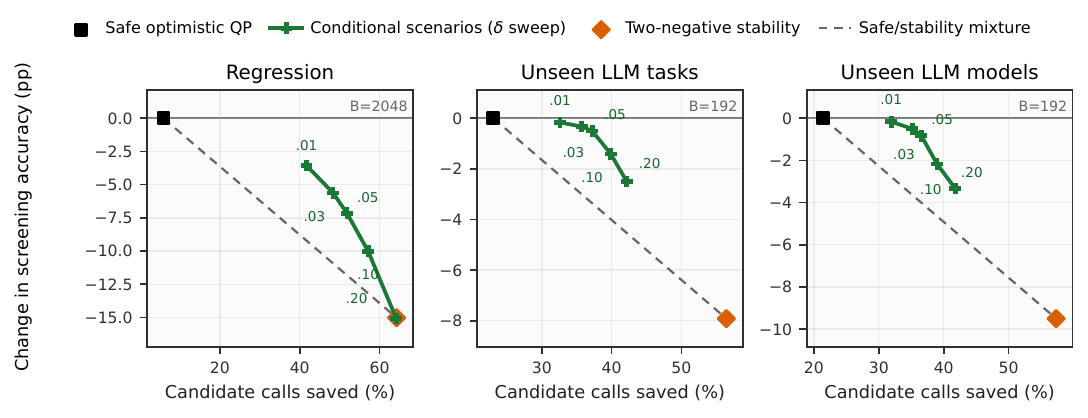}
    \caption{\textbf{Largest-cap posterior threshold sweep.}  Each panel uses
    the same frozen alignment-stratified paths. Both axes compare against
    capped no stopping: call savings and screening accuracy changes.
    Black squares are the safe QP,
    green connected points are conditional posterior scenarios for
    $\delta\in\{0.01,0.03,0.05,0.10,0.20\}$, and orange diamonds are
    two-negative stability. Higher and farther right is better. The gray
    dashed segment joins the expected outcomes of randomized mixtures of the
    two endpoints. The scenario points are model-based and are not certified;
    path repeats are repeated measurements rather than independent cases.}
    \label{fig:posterior-scenario-frontier}
\end{figure}

Table~\ref{tab:posterior-scenario-full-delta} gives the complete threshold
sweep at each panel's largest cap. Figure~\ref{fig:posterior-scenario-frontier}
then visualizes its matched-cap call savings and changes in screening accuracy.
These supplementary plots isolate stopping relative to the same capped
no-stop procedure, using Equations~\eqref{eq:relative-saving}
and~\eqref{eq:screening-agreement-change}.

\begin{table}[!b]
    \centering
    \caption{Full conditional-scenario sweep at each panel's largest cap.
    Savings and agreement changes are relative to capped no stopping;
    $\Delta$ held-out scores usefulness-label accuracy. ``New false rejects'' counts
    paths where full evaluation and capped no stopping keep the candidate but
    scenarios reject it; these are additional final false rejections, not all
    strict early rejections. Denominators are 14,000, 1,200, and 600 paths.
    At $\delta=.05$, regression has 1,008 new final false rejections but
    1,016 strict early false rejections. The values of $\delta$ were swept
    post hoc and carry model-based, pointwise semantics.}
    \label{tab:posterior-scenario-full-delta}
    \setlength{\tabcolsep}{4.7pt}
    \begin{tabular}{@{}lrrrrr@{}}
        \toprule
        Panel & $\delta$ & \shortstack{Calls saved vs.\\capped no-stop (\%)} & \shortstack{$\Delta$ screening\\accuracy (pp)}
              & $\Delta$ held-out (pp) & New false rejects \\
        \midrule
        Regression & .01 & 41.70 & $-3.56$ & $-1.91$ & 503 \\
                   & .03 & 48.36 & $-5.63$ & $-3.21$ & 794 \\
                   & .05 & 51.82 & $-7.16$ & $-4.17$ & 1,008 \\
                   & .10 & 57.13 & $-10.01$ & $-5.90$ & 1,408 \\
                   & .20 & 64.16 & $-15.09$ & $-8.67$ & 2,119 \\
        \addlinespace
        LLM tasks  & .01 & 32.61 & $-0.17$ & $+0.00$ & 2 \\
                   & .03 & 35.72 & $-0.33$ & $+0.17$ & 4 \\
                   & .05 & 37.21 & $-0.50$ & $+0.00$ & 6 \\
                   & .10 & 39.89 & $-1.42$ & $-0.08$ & 17 \\
                   & .20 & 42.17 & $-2.50$ & $-0.83$ & 30 \\
        \addlinespace
        LLM models & .01 & 31.92 & $-0.17$ & $-0.17$ & 1 \\
                   & .03 & 35.17 & $-0.50$ & $-0.50$ & 3 \\
                   & .05 & 36.56 & $-0.83$ & $-0.17$ & 5 \\
                   & .10 & 38.97 & $-2.17$ & $-0.50$ & 13 \\
                   & .20 & 41.78 & $-3.33$ & $-0.33$ & 20 \\
        \bottomrule
    \end{tabular}
\end{table}

Combining Figure~\ref{fig:posterior-scenario-frontier} with the held-out column
of Table~\ref{tab:posterior-scenario-full-delta}, all five LLM threshold points
lie above the matched safe--stability mixture in both screening accuracy and
held-out usefulness-label accuracy. Four of five regression points do so; $\delta=0.20$ falls
just below the mixture. This is an empirical frontier on frozen paths, not
posterior-calibration evidence.

The literal marginal-box diagnostic instead puts a central posterior interval
around each unseen coordinate and gives the optimistic QP the Cartesian product
of those intervals. At $\delta=0.05$, matched-cap savings are 6.32\%, 23.08\%, and
21.42\% in the three panels, compared with 5.89\%, 23.03\%, and 21.42\% for
the safe box; neither policy changes agreement relative to capped no stopping. The intervals remain
wide: their average width is 4.627 on the eight-unit regression support and
0.842 on the unit LLM support, and 45.7\% of LLM role--stratum intervals remain
the complete $[0,1]$ box. Moreover, the optimistic QP may combine all
individually favorable endpoints into a jointly unlikely completion. The
comparison is not a one-factor ablation: the marginal and scenario methods also
use different posterior parameterizations. Their numerical $\delta$ values
therefore should not be interpreted as equivalent guarantees.

\end{document}